\documentclass[sigconf,nonacm,pbalance]{acmart}

\usepackage{bm}
\usepackage{cleveref}
\crefname{figure}{Fig.}{Figs.}
\Crefname{figure}{Fig.}{Figs.}
\crefname{table}{Table}{Tables}
\Crefname{table}{Table}{Tables}
\crefname{section}{Sec.}{Secs.}
\Crefname{section}{Sec.}{Secs.}
\crefname{subsection}{Sec.}{Secs.}
\Crefname{subsection}{Sec.}{Secs.}
\crefname{equation}{Eq.}{Eqs.}
\Crefname{equation}{Eq.}{Eqs.}
\crefname{appendix}{Appendix}{Appendices}
\Crefname{appendix}{Appendix}{Appendices}
\crefname{subappendix}{Appendix}{Appendices}
\Crefname{subappendix}{Appendix}{Appendices}

\newcommand{\testname}{Rubber Ruler Test}
\newcommand{\indep}{\mathrel{\perp\!\!\!\perp}}
\newcommand{\nindep}{\mathrel{\not\!\perp\!\!\!\perp}}

\usepackage{booktabs}
\usepackage{multirow}
\usepackage{tabularx}
\usepackage{makecell}

\usepackage{subcaption}
\usepackage{enumitem}
\newlength{\figh}
\newsavebox{\figboxa}
\newsavebox{\figboxb}
\newsavebox{\figboxc}

\usepackage{placeins}
\providecommand{\Dspline}{\mathcal{D}_{\mathrm{spline}}}

\providecommand{\Ptest}{\mathcal{P}_{\mathrm{test}}}

\providecommand{\indep}{\mathrel{\perp\!\!\!\perp}}
\providecommand{\nindep}{\mathrel{\not\!\perp\!\!\!\perp}}

\crefname{appendix}{Appendix}{Appendices}\Crefname{appendix}{Appendix}{Appendices}
\crefname{subappendix}{Appendix}{Appendices}\Crefname{subappendix}{Appendix}{Appendices}

\begin{document}


\title{Shortcut Learning in Feature-Based Algorithm Selection}


\author{Shuhei Tanaka}
\orcid{0009-0006-3807-4437}
\email{32556009@fukuchiyama.ac.jp}
\affiliation{%
  \institution{The University of Fukuchiyama}
  \city{Fukuchiyama}
  \state{Kyoto}
  \country{Japan}
}

\author{Shoichiro Tanaka}
\orcid{0009-0002-9230-445X}
\email{tanaka-shoichiro@fukuchiyama.ac.jp}
\affiliation{%
  \institution{The University of Fukuchiyama}
  \city{Fukuchiyama}
  \state{Kyoto}
  \country{Japan}
}

\author{Toshiharu Hatanaka}
\orcid{0000-0003-3670-6825}
\email{hatanakatoshiharu@fukuchiyama.ac.jp}
\affiliation{%
  \institution{The University of Fukuchiyama}
  \city{Fukuchiyama}
  \state{Kyoto}
  \country{Japan}
}

\renewcommand{\shortauthors}{Tanaka et al.}

\begin{abstract}
Landscape features give learning models information about the objective function for algorithm selection, algorithm configuration, and performance prediction.
Should these features capture more information?
Not always, because a model may rely on information that the algorithms ignore.
This phenomenon is called shortcut learning, and in-distribution accuracy can hide it.
We propose the \testname{}.
It transforms the objective values of test instances with random strictly increasing transformations, reruns the algorithms, and measures prediction performance.
Comparison-based algorithms such as CMA-ES perform identically after the transformation, so for them any drop reveals shortcut learning.
On 10-dimensional BBOB, a model for two comparison-based algorithms learned shortcuts even from normalized objective values.
The $R^2$ of a model using all ELA features dropped from $0.85$ to $0.23$, below $0.45$ with invariant features only.
Rank-based features reduced the drop when algorithms search using ranks.
Data augmentation with transformed training instances improved prediction on transformed instances for every portfolio we studied, but lowered $R^2$ on original instances by up to 0.31.
These results suggest that models should rely only on information that the algorithms use.
We recommend reporting the \testname{} alongside in-distribution accuracy and using it to choose feature extraction and data augmentation.
\end{abstract}



\begin{teaserfigure}
 \centering
 \includegraphics[width=0.95\textwidth]{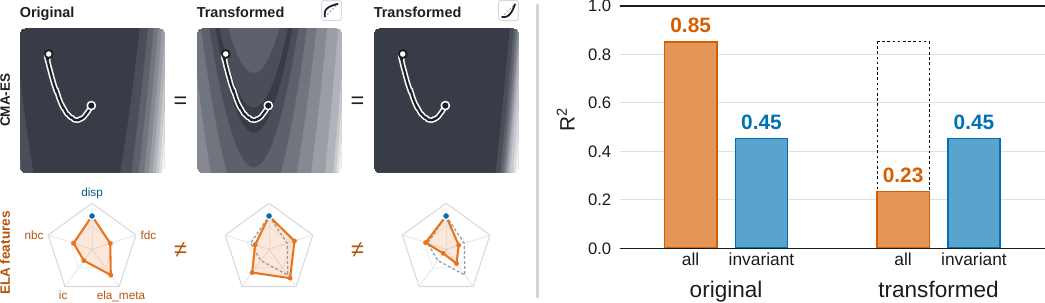}
 \Description{Left: contour plots of an original objective function and two strictly increasing transformations of it. CMA-ES follows the same search path on all three, but radar charts of the ELA feature groups differ across them. Right: a bar chart of $R^2$. The model trained on all ELA features scores 0.85 on the original test instances and 0.23 on the transformed ones; the model trained on invariant features scores 0.45 on both.}
 \caption{
 Even ELA features computed from normalized objective values give a model a shortcut for predicting algorithm performance. A strictly increasing transformation of the objective values does not change the search behavior of a comparison-based algorithm such as CMA-ES (top left), but changes the ELA feature values (bottom left). The model trained on all ELA features (\textit{all}) scores $R^2=0.85$ on the original test instances but only $0.23$ on the transformed ones (right), although the target is the same. The model trained only on features invariant to the transformation (\textit{invariant}, \cref{subsec:exp-features}) keeps $R^2=0.45$ on both test sets. Both models regress the probability that BIPOP CMA-ES beats Adaptive Nelder--Mead on 10-dimensional BBOB (\cref{sec:rq1}).
 }
  \label{fig:teaser}
\end{teaserfigure}

\maketitle

\begin{figure*}[t]
  \centering
  \includegraphics[width=0.95\textwidth]{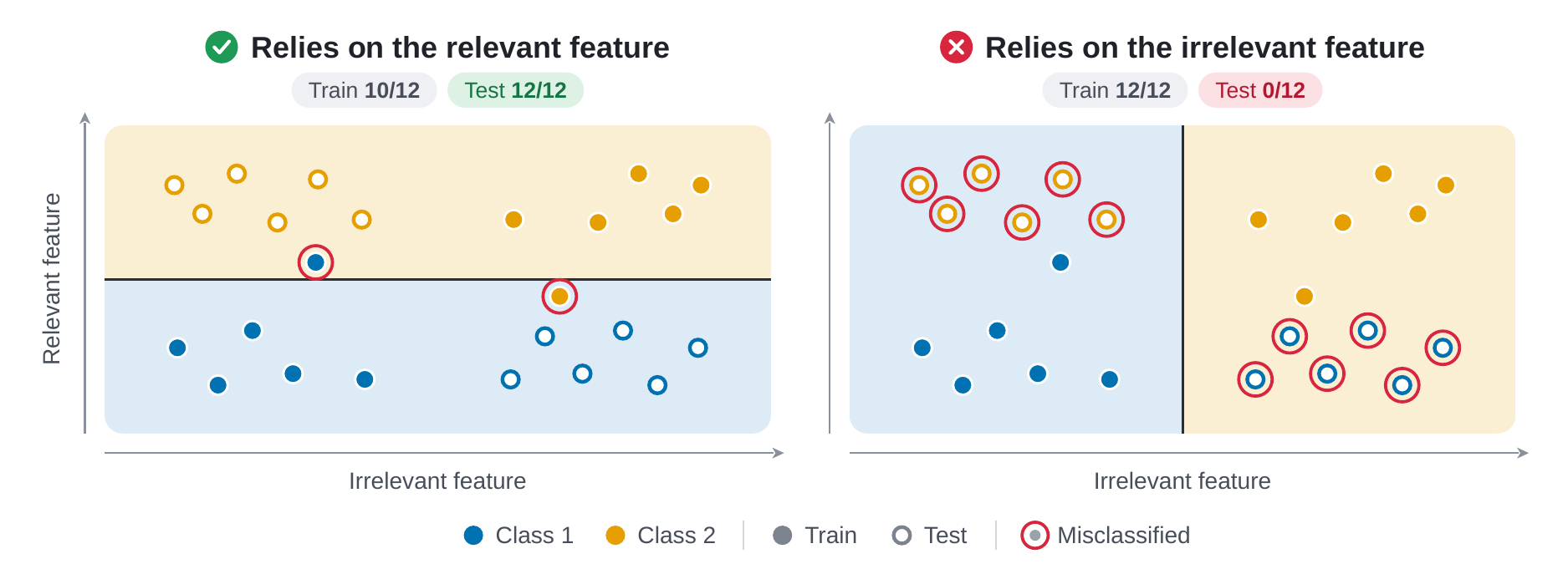}
  \caption{
  A model that relies on an irrelevant feature fails on out-of-distribution test data. The horizontal axis is the irrelevant feature, and the vertical axis is the relevant feature. Filled markers are training data, and hollow markers are test data. In the training data, both features correlate with the class. In the test data, the irrelevant feature's correlation is reversed. The background color shows the decision boundary of each model. The left model misclassifies two training points but correctly classifies all test points. The right model classifies all training points correctly but misclassifies every test point.
  }
  \Description{Two scatter plots of two classes with the irrelevant feature on the horizontal axis and the relevant feature on the vertical axis. The left model relies on the relevant feature and classifies all test points correctly. The right model relies on the irrelevant feature, classifies all training points correctly, and misclassifies every test point.}
  \label{fig:shortcut_learning}
\end{figure*}

\section{Introduction}
Automated algorithm selection (AAS) and automated algorithm configuration (AAC) are promising approaches to choosing an algorithm and its configuration for each black-box optimization problem~\cite{kerschke2019survey}.
This choice matters because the best algorithm and its configuration differ from one optimization problem to another.

Most AAS and AAC methods work in two steps.
First, landscape analysis computes a feature vector for each optimization problem, called a problem instance, from sampled solutions and their objective values.
The most common features come from exploratory landscape analysis (ELA)~\cite{mersmann2011exploratory}.
Second, a trained model maps the feature vector to an algorithm or a configuration, either directly or by predicting algorithm performance.

A common approach to improving AAS and AAC is to design landscape features that capture more information about the objective function~\cite{kerschke2019survey,seiler2025deepela}.
Should they really capture more information?

The answer is not always yes, because a model may rely on information that is irrelevant to the task.
In machine learning, this phenomenon is called shortcut learning~\cite{geirhos2020shortcut}.
Such information can still correlate with the labels in the training data, so the model may learn it as a cue.
As \cref{fig:shortcut_learning} shows, the model then performs well on in-distribution data but poorly on out-of-distribution data.
To trust a model, we must suppress shortcut learning.

AAS and AAC research already applies one remedy: it normalizes the objective values of sampled solutions.
The search behavior of many black-box optimization algorithms is invariant to positive affine transformations of the objective values~\cite{hansen2011impacts}.
A positive multiplier and an additive constant on the objective values are therefore often irrelevant to the task.
Normalization removes this irrelevant information from the model's input.
Prior work has shown that normalization improves the generalization performance of AAS~\cite{prager2023nullifying}.
This improvement can be interpreted as the effect of suppressing shortcut learning.

However, normalization alone is not enough when the portfolio consists of comparison-based algorithms.
Comparison-based algorithms, such as CMA-ES, search for solutions using only the order of objective values~\cite{hansen2001completely}.
Their search behavior is therefore invariant to strictly increasing transformations of the objective values~\cite{hansen2016cma}.
Normalization makes the model's input invariant only to positive affine transformations.
As the left panel of \cref{fig:teaser} shows, a strictly increasing transformation changes some ELA feature values without changing the search behavior of CMA-ES.
Thus, shortcut learning can still occur for comparison-based algorithms.

To investigate shortcut learning in AAS and AAC, we propose the \testname{} (\cref{sec:testname}).
The test trains a model on the original training instances.
It then applies strictly increasing transformations to the objective values of the test instances.
It reruns the algorithms on the transformed instances.
Finally, it measures the prediction performance of the model on the transformed instances.
Measuring a landscape with a rubber ruler does not change the search behavior of comparison-based algorithms.
For such algorithms, a model should therefore output the same predictions.

With the \testname{}, we investigate shortcut learning and ways to suppress it in feature-based AAS on the BBOB (Black-Box Optimization Benchmarking) test suite~\cite{hansen2009real,hansen2021coco}.
We select algorithms from three groups, defined by how they use objective values.
\text{G1} algorithms search using only the order of objective values.
\text{G2} algorithms search using the order and adapt their parameters using the values.
\text{G3} algorithms search using only the values.

Our first claim is that models can still learn shortcuts for a \text{G1} portfolio even with normalized objective values.
As the right panel of \cref{fig:teaser} shows, a model using all ELA features computed from normalized objective values achieved an $R^2$ of $0.85$ on the original test instances.
On the \testname{}, its $R^2$ dropped to $0.23$.
This is below $0.45$, the $R^2$ of a model using only features invariant to the transformation.
Part of the in-distribution $R^2$ of $0.85$ therefore comes from relying on information that the algorithms ignore.

Our second claim is that rank-based feature extraction reduces the drop in prediction performance for \text{G1} and \text{G2} portfolios.
This extraction computes features from the ranks of objective values.
Its $R^2$ stayed at $0.66$ for \text{G1} and changed only from $0.62$ to $0.61$ for \text{G2}.
In contrast, the $R^2$ of value-based extraction dropped to at most~$0.23$.
Adding the value-based features to the rank-based ones did not reduce the drop (at most $0.25$; \cref{fig:rq2-r2}).

Our third claim is that data augmentation with transformed training instances improves prediction performance on transformed test instances for every portfolio we studied.
For \text{G1} and \text{G2}, data augmentation raised the $R^2$ of value-based extraction and concatenation from at most $0.28$ to at least $0.53$.
For \text{G3}, it raised their $R^2$ from at most $-0.81$ to at least $0.13$ on average (\cref{fig:rq3-r2}).

These results suggest an answer to our question.
Models should rely only on information that the algorithms use.
For comparison-based algorithms, rank-based extraction achieves this by removing the other information from the input.
Data augmentation appears to reduce the reliance on such information during training.
In-distribution evaluation ranked the feature extractions differently, with rank-based extraction last for every portfolio.
We therefore recommend that researchers report the \testname{} alongside in-distribution accuracy.
The test applies to any AAS, AAC, or performance prediction model that receives information about the objective function.
Its results also help researchers choose the feature extraction and whether to augment the training data.

Our contributions are threefold:
\begin{itemize}
\item The \testname{}, a test of whether the prediction performance of a model holds under strictly increasing transformations of the objective values (\cref{sec:testname}).
\item Evidence that shortcut learning occurs even with normalized objective values (\cref{sec:rq1}).
\item Two remedies and when each applies: rank-based feature extraction for algorithms that search using ranks (\cref{sec:rq2}) and data augmentation for portfolios that mix them with algorithms that search using values (\cref{sec:rq3}).
\end{itemize}

\section{Related Work}
\label{sec:related}
\v{S}kvorc et al.\ showed that positive affine transformations change many ELA feature values~\cite{skvorc2020understanding}.
Prager and Trautmann argued that features that are not invariant to these transformations can degrade the generalization performance of models on unseen problem instances~\cite{prager2023nullifying}.
They therefore recommended normalizing objective values before computing features.

Petelin and Cenikj showed that leave-instance-out (LIO) evaluation overestimates AAS performance~\cite{petelin2025pitfalls}.
LIO evaluation splits the instances of each BBOB function into training and test data.
They pointed out that LIO evaluation measures how well a model identifies functions, not AAS performance.
They also warned against adopting LIO evaluation as a standard.

This work is novel in two respects.
First, we define which transformations features should be invariant to: those that do not change the algorithms' relative performance, not only positive affine transformations.
Second, we show that part of the AAS performance under LIO evaluation can come from shortcut learning, in which a model relies on irrelevant information.

\section{Preliminaries}
\subsection{Black-Box Optimization}
\label{subsec:def-bbo}
In black-box optimization, no analytical information about the objective function is available.
An optimization algorithm can use only solutions and their objective values.
This work considers single-objective continuous optimization problems with box constraints.
Let $\mathcal{P}$ be a set of problem instances.
We define each problem instance $p\in\mathcal{P}$ as a pair of a feasible region $\mathcal{X}_{p}$ and an objective function $f_{p}$:
\begin{equation}
    p := (\mathcal{X}_{p},\, f_{p}),
    \qquad
    f_{p}\colon\mathcal{X}_{p}\to\mathbb{R},
    \label{eq:instance}
\end{equation}
where $\mathcal{X}_{p}\subset\mathbb{R}^{d}$ is the $d$-dimensional hyperrectangle defined by the box constraints.
The goal of optimization is to find a solution with the smallest possible objective value.

\subsection{Landscape Analysis}
\label{subsec:def-la}
Landscape analysis computes a feature vector for a problem instance from sampled solutions and their objective values.
In this work, landscape analysis also includes sampling the solutions.
Let $\omega\in\Omega$ be the random seed used in landscape analysis.
We define landscape analysis as a mapping
\begin{equation}
    \operatorname{analyzer}\colon\mathcal{P}\times\Omega\to\Phi,
    \label{eq:analyzer}
\end{equation}
where $\Phi$ is the feature space.
We write the feature vector of problem instance $p$ obtained with seed $\omega$ as $\bm{\phi}_{p,\omega}:=\operatorname{analyzer}(p,\omega)$.

\subsection{Automated Algorithm Selection}
\label{subsec:def-aas}
AAS selects a suitable algorithm for each problem instance to obtain better optimization results.
It selects from a set of candidate algorithms $\mathcal{A}$.
In this work, $\mathcal{A}$ is a finite set of two or more algorithms, and we call it the algorithm portfolio.
A selector $\operatorname{selector}\colon\Phi\to\mathcal{A}$ selects an algorithm from a feature vector.
We define AAS as the composition of landscape analysis and the selector:
\begin{equation}
    \operatorname{AAS} := \operatorname{selector}\circ\operatorname{analyzer}
    \colon\mathcal{P}\times\Omega\to\mathcal{A}.
    \label{eq:aas}
\end{equation}

We measure algorithm performance by the mean probability of superiority.
Let $\xi\in\Xi$ be the random seed of an algorithm run.
We regard each algorithm $a\in\mathcal{A}$ as a mapping $a\colon\mathcal{P}\times\Xi\to\mathbb{R}$.
This mapping takes a problem instance $p$ and a random seed $\xi$ and returns the smallest objective value found within a fixed evaluation budget.
Using independent random seeds $\xi$ and $\xi'$, we define the probability of superiority~\cite{vargha2000critique} of $a$ over $a'$ as
\begin{equation}
\begin{aligned}
    \operatorname{PS}(p,a,a') := {}&\Pr\bigl(a(p,\xi)<a'(p,\xi')\bigr)\\
    &+\tfrac{1}{2}\Pr\bigl(a(p,\xi)=a'(p,\xi')\bigr).
\end{aligned}
\label{eq:superiority}
\end{equation}
We then define the mean probability of superiority of $a$ as
\begin{equation}
    J(p,a) := \frac{1}{|\mathcal{A}|-1}\sum_{a'\in\mathcal{A}\setminus\{a\}}\operatorname{PS}(p,a,a').
    \label{eq:mean-ps}
\end{equation}
If each algorithm runs once independently, $J(p,a)$ is the expected fraction of the other algorithms that $a$ beats, counting ties as $1/2$.

For each problem instance $p$, AAS should select an algorithm $a^{*}(p) \in \operatorname*{arg\,max}_{a\in\mathcal{A}} J(p,a)$, that is, an algorithm that beats the most other algorithms on average.

The selector predicts the mean probability of superiority of each algorithm by regression and selects the algorithm with the largest predicted value.
For each algorithm $a\in\mathcal{A}$, a predictor $\operatorname{predictor}_a\colon\Phi\to[0,1]$ maps a feature vector $\bm{\phi}_{p,\omega}$ to a prediction of $J(p,a)$.
We define the selector, with ties broken arbitrarily, as
\begin{equation}
    \operatorname{selector}(\bm{\phi})
    :=
    \operatorname*{arg\,max}_{a\in\mathcal{A}} \operatorname{predictor}_a(\bm{\phi}).
    \label{eq:selector}
\end{equation}

We use regression because it takes into account how much a wrong selection loses, $J(p,a^{*}(p))-J(p,a)$.
Classification with only $a^{*}(p)$ as the label treats all wrong selections equally.

\subsection{Shortcut Learning}
\label{subsec:def-shortcut}
Shortcut learning occurs when a predictor relies on a part of its input that helps prediction only during training~\cite{geirhos2020shortcut}.
We define a predictor $h\colon\mathcal{Z}\to\mathbb{R}$ as a function that predicts a target $y\in\mathbb{R}$ from an input $\bm{z}\in\mathcal{Z}$.
We assume that the pair $(\bm{z},y)$ follows a distribution $\mathcal{D}_{\mathrm{train}}$ during training and a distribution $\mathcal{D}_{\mathrm{deploy}}$ during deployment.
For the predictors in AAS (\cref{eq:selector}), $h$, $\bm{z}$, and $y$ correspond to $\operatorname{predictor}_{a}$, $\bm{\phi}_{p,\omega}$, and $J(p,a)$, respectively.
We split the input $\bm{z}$ into two parts, $\bm{z}=(\bm{z}_{\mathrm{core}},\bm{z}_{\mathrm{spurious}})$.
We write $\indep_{\mathcal{D}}$ for conditional independence under a distribution $\mathcal{D}$ and $\nindep_{\mathcal{D}}$ for its negation.
If $\bm{z}_{\mathrm{core}}$ and $\bm{z}_{\mathrm{spurious}}$ satisfy both
\begin{equation}
\begin{aligned}
    &\bm{z}_{\mathrm{spurious}}\nindep_{\mathcal{D}_{\mathrm{train}}} y \mid \bm{z}_{\mathrm{core}},\\
    &\bm{z}_{\mathrm{spurious}}\indep_{\mathcal{D}_{\mathrm{deploy}}} y \mid \bm{z}_{\mathrm{core}},
\end{aligned}
\label{eq:shortcut-def}
\end{equation}
we call $\bm{z}_{\mathrm{core}}$ the core feature and $\bm{z}_{\mathrm{spurious}}$ the spurious feature.
In other words, during training, the spurious feature carries information about $y$ that the core feature lacks.
During deployment, it carries no such information.
With these terms, shortcut learning means that a predictor relies on the spurious feature.

\section{\testname}
\label{sec:testname}
The \testname{} checks whether the prediction performance of a trained model holds when a strictly increasing transformation is applied to the objective values.
The user provides the elements in the upper part of \cref{tab:testname-elements}; the test fixes the rest.

\subsection{Extracting Features}
\label{subsec:extraction}
The \testname{} compares value-based and rank-based feature extraction.
Both pass information about the objective values to the user-provided landscape analysis $\operatorname{analyzer}$.
Value-based feature extraction $\operatorname{analyzer}_{\mathrm{value}}$ passes the min-max normalized objective values of the sampled solutions.
Rank-based feature extraction $\operatorname{analyzer}_{\mathrm{rank}}$ passes the min-max normalized ranks of these objective values and assigns average ranks to ties.
For $\tau \in \{\mathrm{value}, \mathrm{rank}\}$, we write the feature vector as $\bm{\phi}^{\tau}_{p,\omega} := \operatorname{analyzer}_{\tau}(p, \omega)$.

The two feature extractions differ in their invariance.
Because of min-max normalization, value-based features are invariant to positive affine transformations but generally not to strictly increasing transformations.
Rank-based features are invariant to both, because neither changes the ranks.
That is, for any strictly increasing $g$ and the same seed $\omega$, $\bm{\phi}^{\mathrm{rank}}_{g\circ p,\omega} = \bm{\phi}^{\mathrm{rank}}_{p,\omega}$, where $g\circ p := (\mathcal{X}_{p}, g\circ f_{p})$.

\subsection{Transforming Test Instances}
\label{subsec:transform}
Let $\psi$ be a strictly increasing function on $[0,1]$ with $\psi(0) = 0$ and $\psi(1) = 1$.
For a problem instance $p$, we define the transformation
\begin{equation}
    g_{\psi,p}(v) := \bar{\psi}\left( \frac{v - \hat{f}^{\min}_{p}}{\hat{f}^{\max}_{p} - \hat{f}^{\min}_{p}} \right),
    \label{eq:warp}
\end{equation}
where $\hat{f}^{\min}_{p}$ and $\hat{f}^{\max}_{p}$ are the minimum and maximum of $f_{p}$ over $1000d$ solutions sampled uniformly at random from $\mathcal{X}_{p}$, or the known values.
The function $\bar{\psi}$ extends $\psi$ outside $[0,1]$ by the identity.
Because $\psi(0) = 0$ and $\psi(1) = 1$, $\bar{\psi}$ is strictly increasing on $\mathbb{R}$, so values outside the estimated range also keep their order.

The \testname{} draws $\psi$ from a distribution $\mathcal{D}_{\mathrm{spline}}$ (examples in \cref{fig:monotone-warp}).
To draw $\psi$, we draw the number of intervals $K$ uniformly from $\{4, \dots, 7\}$ and heights $c_{0}, \dots, c_{K}$ independently from $\mathcal{N}(0, \sigma^{2})$.
Let $s$ be the not-a-knot cubic spline through the points $(k/K, c_{k})$.
Finally, we set
\begin{equation}
    \psi(t) := \frac{\int_{0}^{t} \exp\bigl(s(u)\bigr)\,\mathrm{d}u}{\int_{0}^{1} \exp\bigl(s(u)\bigr)\,\mathrm{d}u},
    \label{eq:spline-transform}
\end{equation}
which is strictly increasing because $\exp\bigl(s(u)\bigr) > 0$.

\subsection{Evaluating Trained Models}
\begin{table}[t]
    \centering
    \caption{Elements that the user provides (upper part) and that the \testname{} fixes (lower part).}
    \label{tab:testname-elements}
    \begin{tabular}{ll}
        \toprule
        Element & Setting \\
        \midrule
        \multicolumn{2}{l}{\textbf{Provided by the user}} \\
        Portfolio & --- \\
        Landscape analysis & --- \\
        Learning model & --- \\
        Training instances & --- \\
        Test instances & --- \\
        \midrule
        \multicolumn{2}{l}{\textbf{Fixed by the test}} \\
        Number of samples for $\hat{f}^{\min}_{p}, \hat{f}^{\max}_{p}$ & $1000d$, or known values \\
        Number of intervals $K$ & Uniform on $\{4, \dots, 7\}$ \\
        Heights $c_k$ & $\mathcal{N}(0, \sigma^2)$, $\sigma = 1$ \\
        Evaluation metric for predictors & $R^{2}$ \\
        Evaluation metric for selectors & $\Theta$ \\
        \bottomrule
    \end{tabular}
\end{table}
\begin{table}[t]
    \centering
    \caption{Compared conditions and portfolios in each RQ. All RQs use ELA features. Mixed is the portfolio $\mathrm{G1} \cup \mathrm{G2} \cup \mathrm{G3}$.}
    \label{tab:rq-design}
    \setlength{\tabcolsep}{3pt}
    \begin{tabular}{@{}llcccc@{}}
        \toprule
        & & \multicolumn{3}{c}{Regression} & Selection \\
        \cmidrule(lr){3-5} \cmidrule(l){6-6}
        RQ & Compared conditions & $\mathrm{G1}$ & $\mathrm{G2}$ & $\mathrm{G3}$ & Mixed \\
        \midrule
        RQ1 & \makecell[l]{All value-based features /\\ invariant features} & \checkmark & & & \\
        \addlinespace
        RQ2 & Three feature extractions & \checkmark & \checkmark & \checkmark & \checkmark \\
        \addlinespace
        RQ3 & \makecell[l]{With / without data augmentation\\ $\times$ three feature extractions} & \checkmark & \checkmark & \checkmark & \checkmark \\
        \bottomrule
    \end{tabular}
\end{table}
\label{subsec:test-procedure}
\label{subsec:benchmark}
The \testname{} uses the disjoint training set $\mathcal{P}_{\mathrm{train}}$ and test set $\mathcal{P}_{\mathrm{test}}$ provided by the user, and the transformed test set
\begin{equation}
    \widetilde{\mathcal{P}}_{\mathrm{test}} := \bigl\{ g_{\psi_{p},p} \circ p \bigm| p \in \mathcal{P}_{\mathrm{test}} \bigr\},
    \quad
    \psi_{p} \overset{\text{i.i.d.}}{\sim} \mathcal{D}_{\mathrm{spline}}.
    \label{eq:transformed-test}
\end{equation}
Each test instance $p$ and its transformed instance $g_{\psi_{p},p} \circ p$ share $\omega_{p}$ and the seeds $\xi$ of all algorithm runs.
For each $q \in \widetilde{\mathcal{P}}_{\mathrm{test}}$, we obtain $J(q, a)$ by running every algorithm on $q$.

For each $\tau$ and $a \in \mathcal{A}$, the \testname{} trains a predictor $\operatorname{predictor}^{\tau}_{a}$ with the same user-provided learning model to predict $J(p, a)$ from $\bm{\phi}^{\tau}_{p,\omega_{p}}$ on $p \in \mathcal{P}_{\mathrm{train}}$.
We write its prediction as $\hat{J}_{\tau}(p, a)$.

For a set of problem instances $\mathcal{Q}$, the \testname{} measures prediction performance by $R^{2}_{\tau}(\mathcal{Q})$.
This is the coefficient of determination between $\hat{J}_{\tau}(q, a)$ and $J(q, a)$ over $q \in \mathcal{Q}$, averaged over $a \in \mathcal{A}$.
For a selector, the \testname{} instead uses $\Theta$, the probability of superiority over the single best solver (SBS)~\cite{kerschke2019survey} (\cref{eq:win-rate-vs-sbs}).
The \testname{} reports $R^{2}_{\tau}(\mathcal{P}_{\mathrm{test}})$ and $R^{2}_{\tau}(\widetilde{\mathcal{P}}_{\mathrm{test}})$ for each $\tau$.
A drop shows that the prediction performance does not hold under the transformation.
If the transformation does not change $J$, as for comparison-based algorithms, a drop reveals shortcut learning.

The \testname{} also outputs
\begin{equation}
    \tau^{*} \in \operatorname*{arg\,max}_{\tau \in \{\mathrm{value}, \mathrm{rank}\}} R^{2}_{\tau}(\widetilde{\mathcal{P}}_{\mathrm{test}}).
    \label{eq:decision}
\end{equation}
Because $\tau^{*}$ is chosen on $\widetilde{\mathcal{P}}_{\mathrm{test}}$, its $R^{2}$ there can be optimistic.
If $R^{2}_{\tau^{*}}(\widetilde{\mathcal{P}}_{\mathrm{test}})$ is still lower than $R^{2}_{\tau^{*}}(\mathcal{P}_{\mathrm{test}})$, the user can consider data augmentation.
The user can also add candidates to \cref{eq:decision}, such as concatenation or data augmentation.

\section{Experimental Setup}
\label{sec:experiments}
\begin{figure*}[t]
  \centering
  \setlength{\figh}{0.248\textwidth}
  \begin{subfigure}[t]{0.765\textwidth}
    \centering
    \includegraphics[height=\figh]{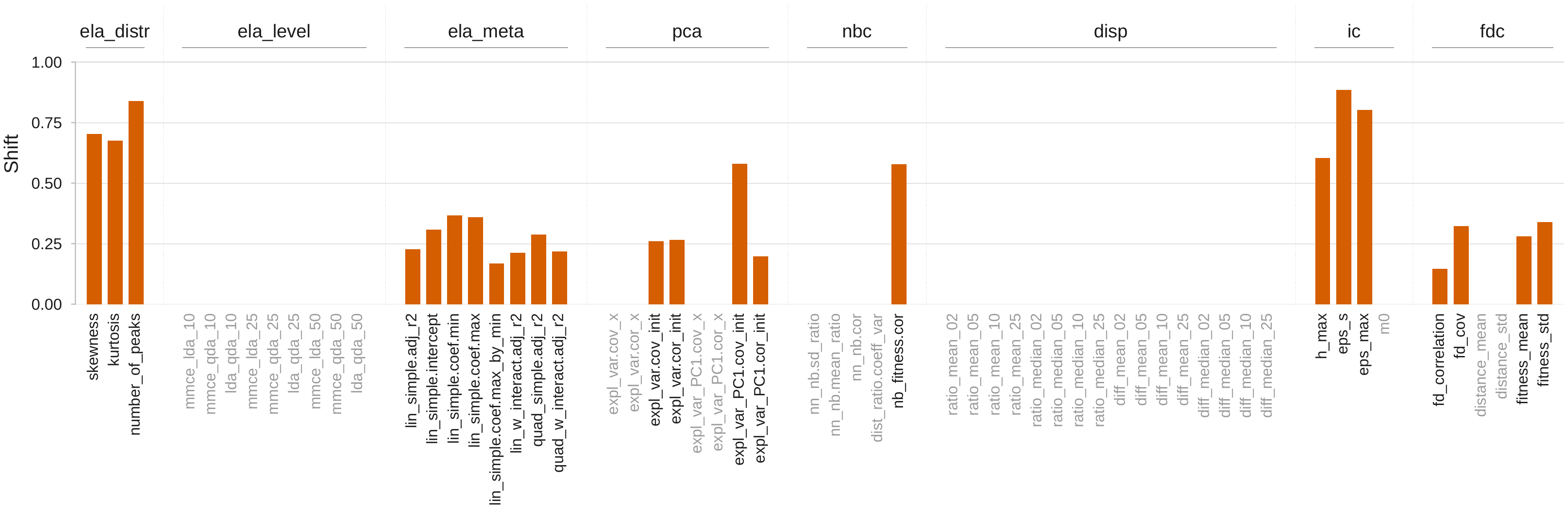}
    \caption{ELA features.}
    \label{fig:pre-variation-ela}
  \end{subfigure}\hfill
  \begin{subfigure}[t]{0.227\textwidth}
    \centering
    \includegraphics[height=\figh]{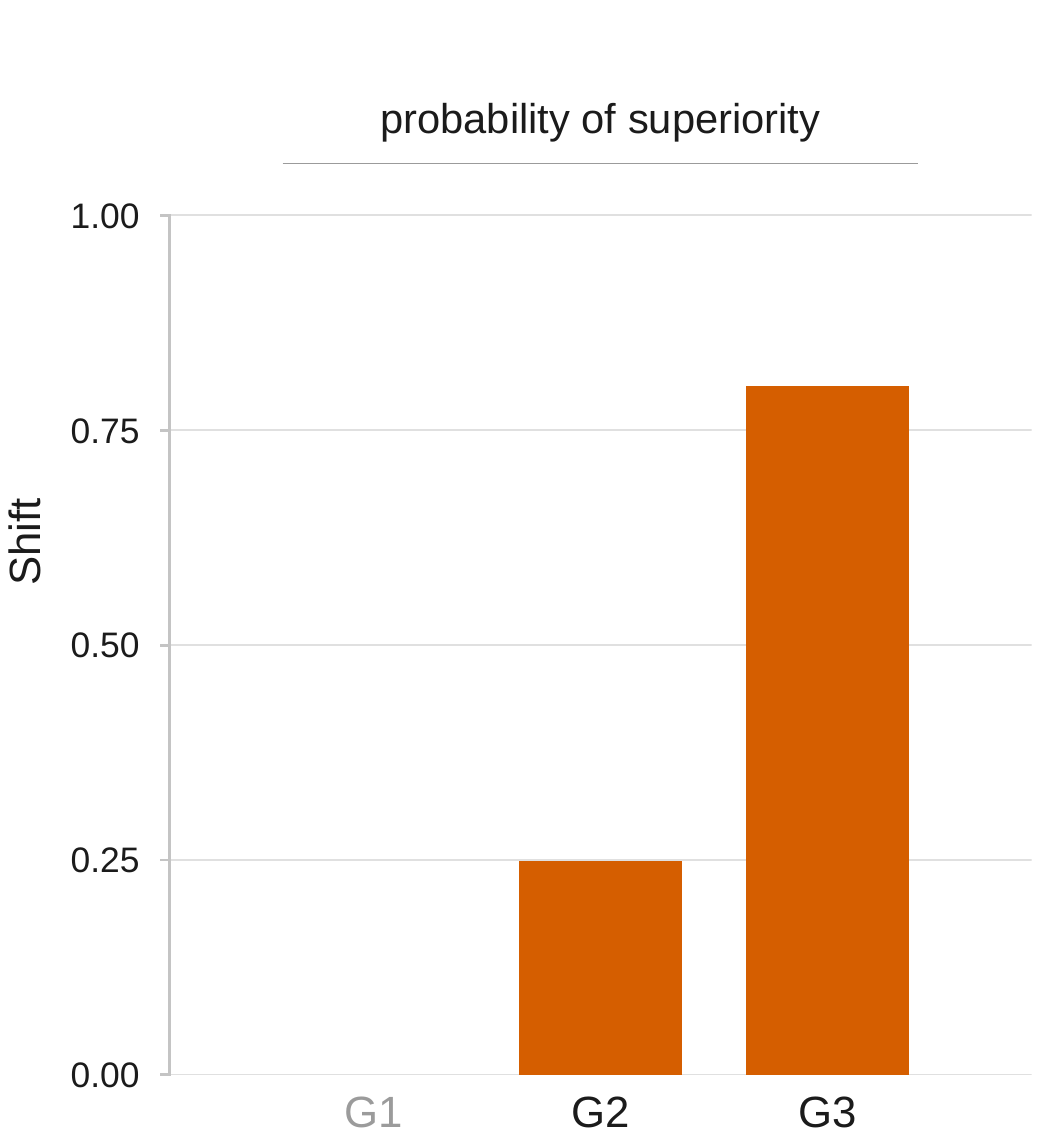}
    \caption{$J$ of each group.}
    \label{fig:pre-variation-target}
  \end{subfigure}
  \caption{Strictly increasing transformations change only non-invariant features, not $J$ of G1. The vertical axis, labeled Shift, is $\Delta_{\eta}$ in \cref{eq:variation}. Gray names are invariant features.}
  \Description{Left: bar chart of the change of each of the 59 ELA features under the transformation; the 36 invariant features have zero change, and the other 23 have a median change of 0.32. Right: bar chart of the change of J for each group; it is 0 for G1, 0.25 for G2, and 0.80 for G3.}
  \label{fig:pre-variation}
\end{figure*}
\subsection{Overview}
\label{subsec:overview}
We investigate three research questions:
\begin{description}
    \item[RQ1] Does shortcut learning occur even with normalized objective values?
    \item[RQ2] Does rank-based feature extraction reduce the drop in prediction performance?
    \item[RQ3] Does data augmentation improve prediction performance on transformed test instances?
\end{description}
\Cref{tab:rq-design} shows the tasks and the conditions compared in each RQ.
In every task, the predictors predict $J$, and we train and evaluate them with the procedure in \cref{subsec:test-procedure}.
All settings and instructions for reproduction are in our repository\footnote{\url{https://github.com/shuheitnk/shortcut-learning-aas}}.
We implemented the code with Claude Code (Anthropic).
We verified it through code review, automated tests, run-time checks such as the invariance of $J$ for G1, and a script that checks the numbers quoted in the paper against the outputs.

We use two tasks: regression and selection.
The regression task uses the G1, G2, and G3 portfolios, each of which consists of two algorithms from the same group (\cref{tab:exp-algorithms}).
For G1, the transformation does not change $J$ (\cref{subsec:exp-portfolio}), so a drop in prediction performance indicates shortcut learning.
For G2 and G3, the transformation also changes $J$, so a drop can also come from failing to track this change.
The selection task uses the mixed portfolio $\mathrm{G1} \cup \mathrm{G2} \cup \mathrm{G3}$ of all six algorithms and measures the performance of AAS by $\Theta$ (\cref{subsec:exp-metrics}).

RQ1 tests whether the min-max normalization in value-based feature extraction (\cref{subsec:extraction}) prevents shortcut learning.
In the G1 regression task, we compare a predictor that uses all 59 value-based ELA features (value-based extraction in RQ2) with a baseline that uses only invariant features (\cref{subsec:exp-features}).
The transformation changes neither the input of the baseline nor $J$.
We conclude that shortcut learning occurs if the predictor with all features has a higher median $R^{2}$ than the baseline before the transformation but a lower median $R^{2}$ after it.
The non-invariant features then carry the spurious feature in \cref{eq:shortcut-def}, mixed with information that stays valid under the transformation (\cref{sec:rq3}).

RQ2 and RQ3 test the two remedies on both tasks.
RQ2 compares three feature extractions: value-based, rank-based, and concatenation ($\tau = \mathrm{concat}$).
Concatenation joins the value-based and rank-based feature vectors. We add it to examine whether more information improves prediction performance.
RQ3 compares each feature extraction with and without data augmentation, which RQ1 and RQ2 do not use.

\subsection{Problem Instances}
\label{subsec:exp-instances}
We create 10-dimensional problem instances from the 24 BBOB functions~\cite{hansen2009real}, with the feasible region $[-5, 5]^{10}$.
The training set $\mathcal{P}_{\mathrm{train}}$ consists of instances 11--110 of each function (2,400 instances).
The test set $\mathcal{P}_{\mathrm{test}}$ consists of instances 1--10 of each function (240 instances).
This split is the LIO evaluation in \cref{sec:related}~\cite{petelin2025pitfalls}.

We create the transformed test set $\widetilde{\mathcal{P}}_{\mathrm{test}}$ with the procedure in \cref{subsec:test-procedure}, using the known minimum of each instance as $\hat{f}^{\min}_{p}$.
RQ1 and RQ2 use $\mathcal{D}_{\mathrm{spline}}$.
RQ3 instead uses four distributions that data augmentation does not use: $\mathcal{D}_{\mathrm{spline}}$ with $K\in\{2,3\}$ or $K\in\{14,\dots,19\}$, power functions $\psi(t)=t^{\gamma}$, and piecewise-linear functions (Appendix~\ref{app:settings}).

\begin{table}[t]
    \centering
    \caption{Algorithms and the information about objective values that they use in their search.}
    \label{tab:exp-algorithms}
    \setlength{\tabcolsep}{4pt}
    \begin{tabular}{@{}lll@{}}
        \toprule
        Group & Algorithm & Information used \\
        \midrule
        G1 & BIPOP CMA-ES~\cite{hansen2009bipop} & Ranks \\
           & Adaptive Nelder--Mead~\cite{gao2012implementing} & Ranks \\
        \midrule
        G2 & SHADE~\cite{tanabe2013success} & Ranks and values \\
           & L-SHADE~\cite{tanabe2014improving} & Ranks and values \\
        \midrule
        G3 & BOBYQA~\cite{powell2009bobyqa} & Values \\
           & Dual Annealing~\cite{xiang1997generalized} & Values \\
        \bottomrule
    \end{tabular}
\end{table}

\subsection{Portfolios and Performance Measurement}
\label{subsec:exp-portfolio}
We use six algorithms, grouped by the information about objective values that they use in their search (\cref{tab:exp-algorithms}).
Each algorithm runs 31 times on each problem instance, and each run uses $1000d$ evaluations, including restarts.
Each algorithm restarts when its normalized search scale falls below a threshold; this rule does not use objective values.
We estimate the probability of superiority in \cref{eq:superiority} from all pairs of runs.
Appendix~\ref{app:settings} lists the parameter settings and how we tuned them.

\subsection{Feature Extraction}
\label{subsec:exp-features}
We compute ELA features with \texttt{pflacco}~\cite{prager2024pflacco} from $50d = 500$ solutions sampled by Latin hypercube sampling~\cite{mckay1979comparison}.
We compute the features once for each training instance, with one seed $\omega$, and 11 times for each test instance, with 11 seeds $\omega$.
We drop the two features that take a non-finite value on some training or test instance, which leaves 59 features for all experiments.
Of the 59 features, 36 depend only on the coordinates of the solutions or on the order of their objective values.
We call them invariant features (Appendix~\ref{app:settings}) because strictly increasing transformations do not change them.

Appendix~\ref{app:deep-ela} shows that the 48 features of the pretrained Deep-ELA~\cite{seiler2025deepela} lead to the same conclusions in RQ2 and RQ3.

\subsection{Learning Model and Data Augmentation}
\label{subsec:exp-model}
As the learning model, we use RealMLP-TD~\cite{holzmuller2024better} from \texttt{pytabkit} with its default settings.
For each condition, we train predictors with 11 random seeds of the learning model.
Appendix~\ref{app:models} shows the results with random forest~\cite{breiman2001random} and XGBoost~\cite{chen2016xgboost}.

Data augmentation adds one transformed instance $g_{\psi_{p},p} \circ p$ with $\psi_{p} \sim \mathcal{D}_{\mathrm{spline}}$ to the training set for each training instance $p$.
Each added instance shares $\omega$ and the seeds $\xi$ with $p$, and we obtain its $J$ by running the algorithms on it.

\subsection{Evaluation Metrics and Aggregation}
\label{subsec:exp-metrics}
For regression, we compute $R^{2}_{\tau}$ on the 240 test instances for each $\omega$.

In the selection task, we compare AAS, which uses the selector in \cref{eq:selector}, with the single best solver (SBS).
The SBS is the single algorithm with the highest mean $J$ on $\mathcal{P}_{\mathrm{train}}$, which is BIPOP CMA-ES.
AAS and the SBS both receive $1050d$ evaluations.
AAS spends $50d$ evaluations on computing features and $1000d$ evaluations on running the selected algorithm, and it returns the best objective value among all its evaluations.
For a set of problem instances $\mathcal{Q}$, we define the selection performance as
\begin{equation}
    \Theta(\mathcal{Q}) := \frac{1}{|\mathcal{Q}|}\sum_{q\in\mathcal{Q}} \operatorname{PS}(q, \mathrm{AAS}, \mathrm{SBS}).
    \label{eq:win-rate-vs-sbs}
\end{equation}
A value of $\Theta > 0.5$ means that AAS beats the SBS on average.
As a reference, we also report $\Theta$ of an Oracle that predicts $J$ perfectly.
Like AAS, the Oracle spends $50d$ evaluations on computing features and $1000d$ on the selected algorithm, so it can lose to the SBS.

Each condition gives one $R^{2}_{\tau}$ and one $\Theta$ for each pair of a seed of the learning model and a seed $\omega$.
For each seed of the learning model, we take the median over the 11 seeds $\omega$.
In RQ3, we then average over the four distributions.
We report the median of the resulting 11 values, and the statistical tests use these values.

\subsection{Statistical Tests}
\label{subsec:exp-tests}
We test the differences on the transformed test instances with two-sided exact tests at $\alpha = 0.05$ with Holm correction~\cite{holm1979simple} within each family of comparisons (Appendix~\ref{app:settings}).
We compare feature extractions with the Wilcoxon signed-rank test~\cite{wilcoxon1945individual}, paired by the seed of the learning model (11 pairs).
We compare training with and without data augmentation with the Mann--Whitney $U$ test~\cite{mann1947test} (11 vs.\ 11 seeds).
For RQ1, we also compare the predictor with all features and the baseline with the Wilcoxon signed-rank test, without correction, to supplement the criterion in \cref{subsec:overview}.

These tests cover only the variation across model seeds; ``significant'' below refers to them.
We note where a bootstrap interval over $\omega$ and the test instances (Appendix~\ref{app:bootstrap}) includes zero.

\section{Results}

\begin{figure*}[t]
  \centering
  \begin{subfigure}[t]{0.48\textwidth}
    \centering
    \includegraphics[width=\linewidth]{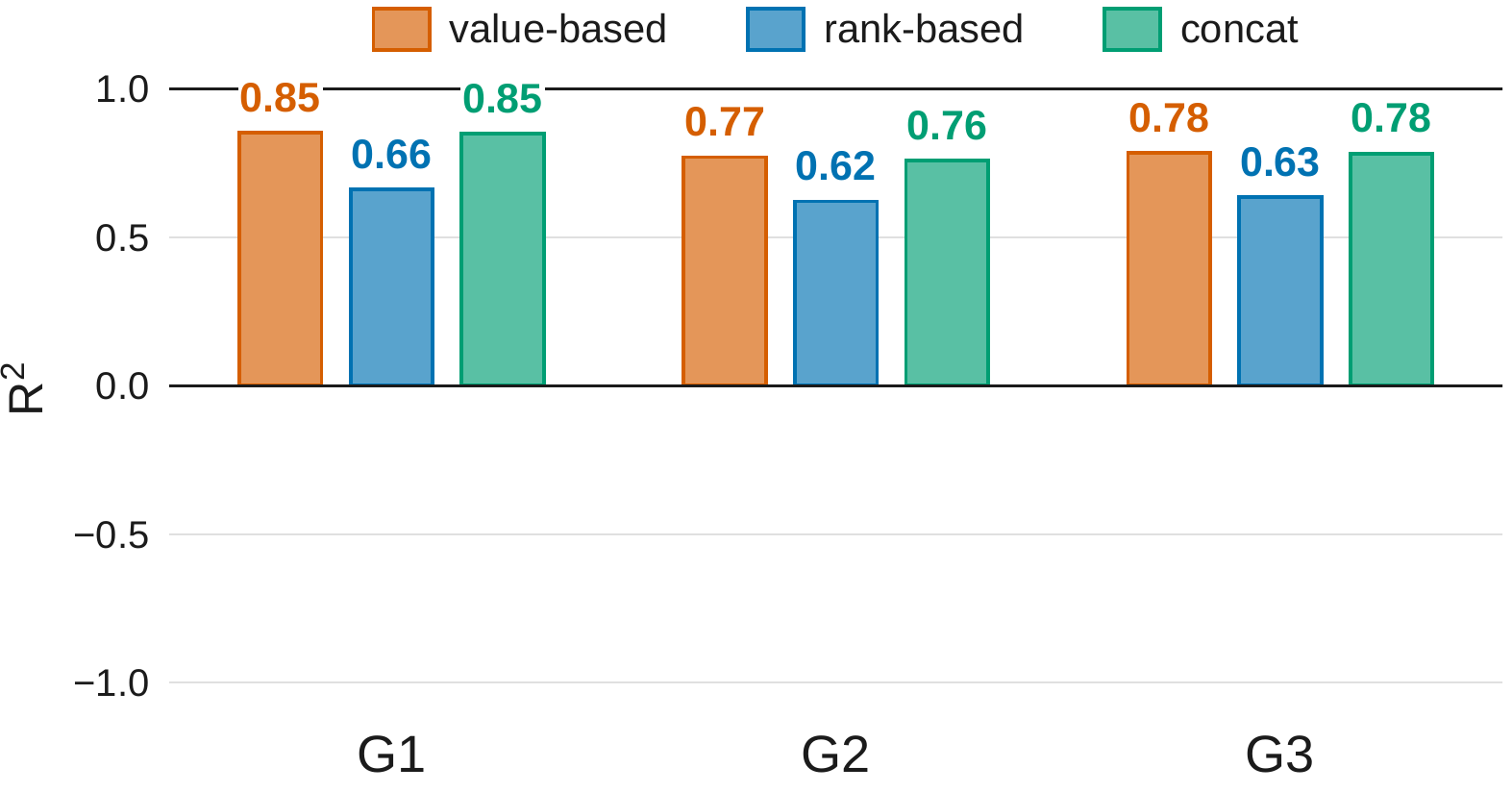}
    \caption{Original test instances.}
    \label{fig:rq2-r2-ela-orig}
  \end{subfigure}\hfill
  \begin{subfigure}[t]{0.48\textwidth}
    \centering
    \includegraphics[width=\linewidth]{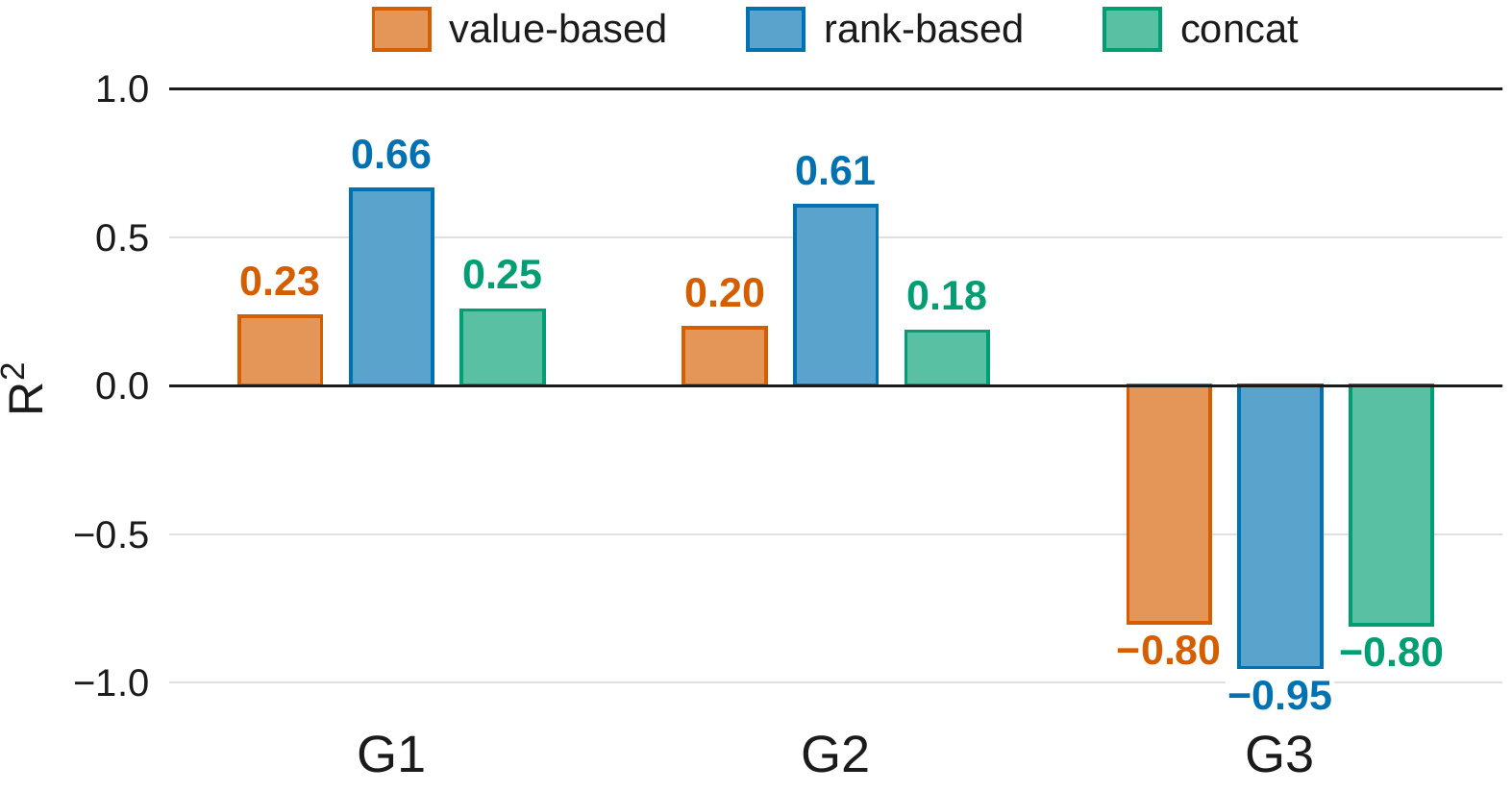}
    \caption{Transformed test instances.}
    \label{fig:rq2-r2-ela-trans}
  \end{subfigure}
  \caption{Rank-based extraction has the lowest $R^{2}$ on original test instances but the highest for G1 and G2 on transformed ones.}
  \Description{Two panels of R-squared values for value-based, rank-based, and concatenated ELA feature extraction on the G1, G2, and G3 portfolios. Left: original test instances; right: transformed test instances. On the original instances, rank-based extraction is the lowest in every group. On the transformed instances, rank-based extraction stays at 0.61 or higher for G1 and G2, whereas the other two drop to 0.25 or lower; all three are negative for G3.}
  \label{fig:rq2-r2}
\end{figure*}
\subsection{Preliminary Analysis}
\label{sec:pre_study}
This analysis checks two preconditions for non-invariant features to act as spurious features.
Strictly increasing transformations should change these features but not $J$ of G1.
We measure the change in a feature or in $J$, denoted by $\eta$, as
\begin{equation}
    \Delta_{\eta} := \frac{1}{s_{\eta}}\operatorname*{mean}_{p\in\mathcal{P}_{\mathrm{test}}}\bigl\lvert \eta(\tilde{p}) - \eta(p) \bigr\rvert,
    \label{eq:variation}
\end{equation}
where $\tilde{p} := g_{\psi_{p},p}\circ p$ and $s_{\eta}$ is the standard deviation of $\eta$ over $\mathcal{P}_{\mathrm{test}}$.
The transformation changed only the 23 non-invariant features, with a median change of 0.32 (\cref{fig:pre-variation-ela}).
$J$ of G1 was identical on all 240 instances, and the change in $J$ was 0.25 for G2 and 0.80 for G3 (\cref{fig:pre-variation-target}).
Both preconditions therefore hold for G1.

\subsection{RQ1: Does Shortcut Learning Occur Even with Normalized Objective Values?}
\label{sec:rq1}
Shortcut learning occurred in the G1 regression task even with normalized objective values (right panel of \cref{fig:teaser}).
Before the transformation, the predictor with all features had an $R^{2}$ of 0.85, higher than the 0.45 of the baseline.
After the transformation, its $R^{2}$ dropped to 0.23, below the baseline, whose $R^{2}$ did not change.
This result meets the criterion in \cref{subsec:overview}, and the difference between the two predictors after the transformation was significant ($p < 0.001$).
With four more draws of $\psi$, this $R^{2}$ was between $-0.04$ and 0.13 (Appendix~\ref{app:controls}).
Random forest and XGBoost agreed (Appendix~\ref{app:models}).

For G1, only the non-invariant features change (\cref{sec:pre_study}).
The drop therefore shows that the predictor relied on them, even though it also received all 36 invariant features.

\subsection{RQ2: Does Rank-Based Feature Extraction Reduce the Drop in Prediction Performance?}
\label{sec:rq2}
Rank-based feature extraction reduced the drop in prediction performance for the G1 and G2 portfolios.
After the transformation, rank-based extraction had an $R^{2}$ of at least 0.61 for G1 and G2, significantly higher than value-based extraction, which had at most 0.23 (\cref{fig:rq2-r2-ela-trans}).
For G1, its $R^{2}$ did not change, because the transformation changes neither its input nor $J$.
For G2, its $R^{2}$ changed by only 0.01, although $J$ changed by 0.25.

Rank-based extraction did not reduce the drop for G3.
After the transformation, all feature extractions had a negative $R^{2}$ for G3 (at most $-0.80$; \cref{fig:rq2-r2-ela-trans}).
The transformation changes $J$ of G3 the most (\cref{fig:pre-variation-target}), because G3 algorithms search using values.
With $\psi$ set to the identity, the rescaling in \cref{eq:warp} alone changed $J$ of G3 by 0.65 (Appendix~\ref{app:controls}).
The acceptance rule of Dual Annealing, unlike BOBYQA, depends on the scale of objective values, which both min-max normalization and ranks remove.
This suggests that the input should also keep what the algorithms use.

The drawback of rank-based extraction is a lower $R^{2}$ before the transformation: 0.15--0.19 lower than that of value-based extraction for every group (\cref{fig:rq2-r2-ela-orig}).
For G1, $J$ depends only on the order of objective values.
The advantage of value-based extraction for G1 therefore comes at least partly from information that is valid on the original test instances but not under the transformation (\cref{sec:rq1}).

\begin{figure}[t]
  \centering
  \includegraphics[width=\columnwidth]{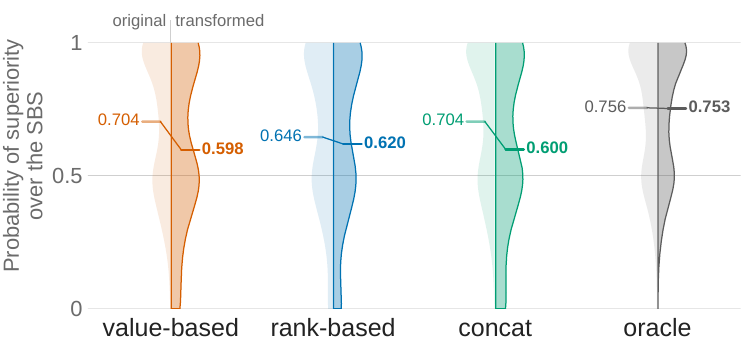}
  \caption{On the transformed test instances, rank-based feature extraction gives a slightly higher $\Theta$ than the others in the selection task.
  In each violin, the left half shows the original test instances and the right half shows the transformed ones.
  The numbers are $\Theta$.}
  \Description{Split violin plots of the probability of superiority of AAS over the SBS across test instances, for value-based, rank-based, and concatenated ELA feature extraction and the Oracle. On the original instances, rank-based extraction has the lowest Theta (0.65, versus 0.70 for the others). On the transformed instances, it has the highest Theta (0.62, versus 0.60 for the others). The Oracle has 0.76 and 0.75.}
  \label{fig:rq2-aas}
\end{figure}

\begin{figure*}[t]
  \centering
  \begin{subfigure}[t]{0.48\textwidth}
    \centering
    \includegraphics[width=\linewidth]{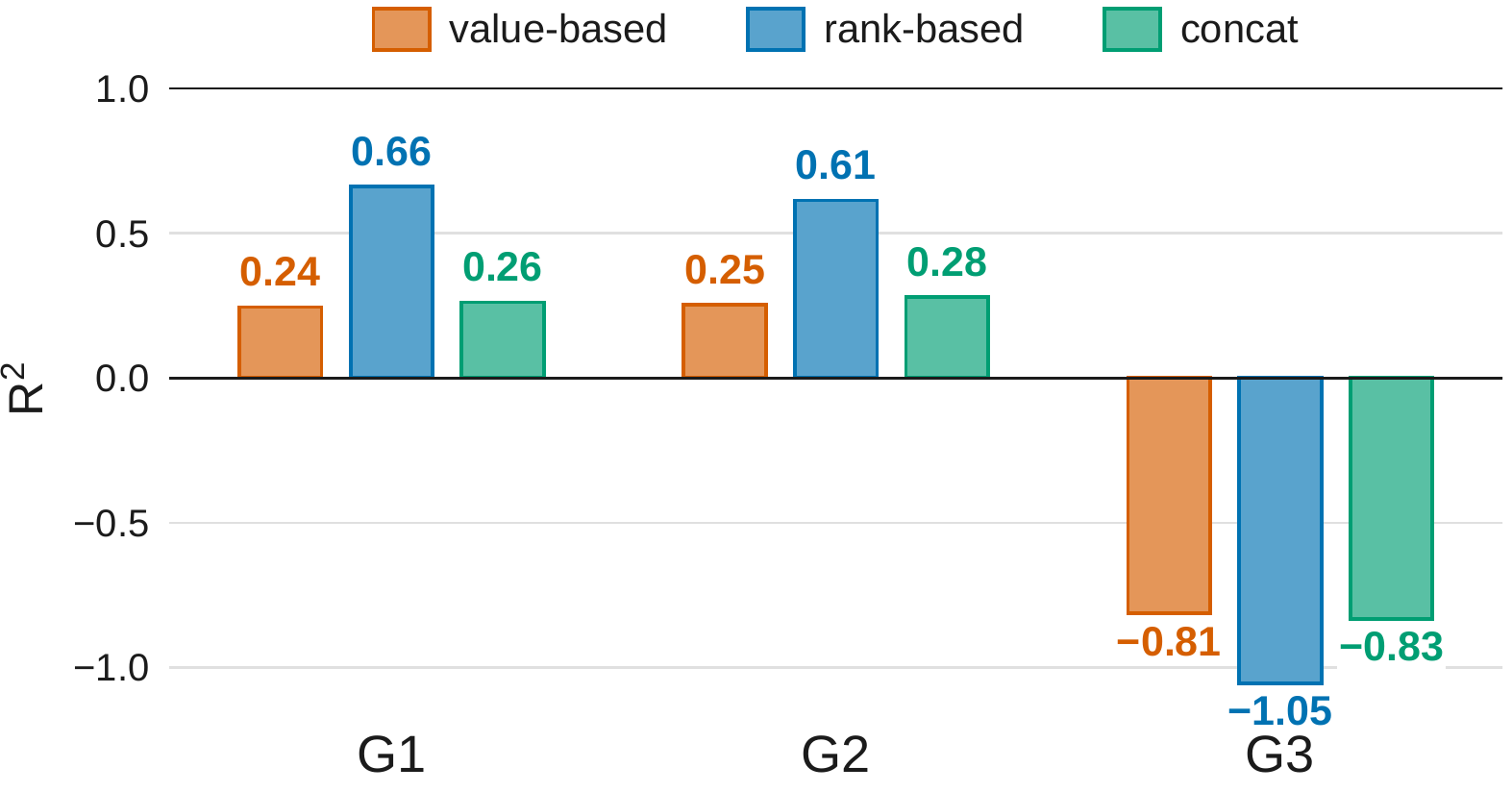}
    \caption{Without data augmentation.}
    \label{fig:rq3-r2-ela-plain}
  \end{subfigure}\hfill
  \begin{subfigure}[t]{0.48\textwidth}
    \centering
    \includegraphics[width=\linewidth]{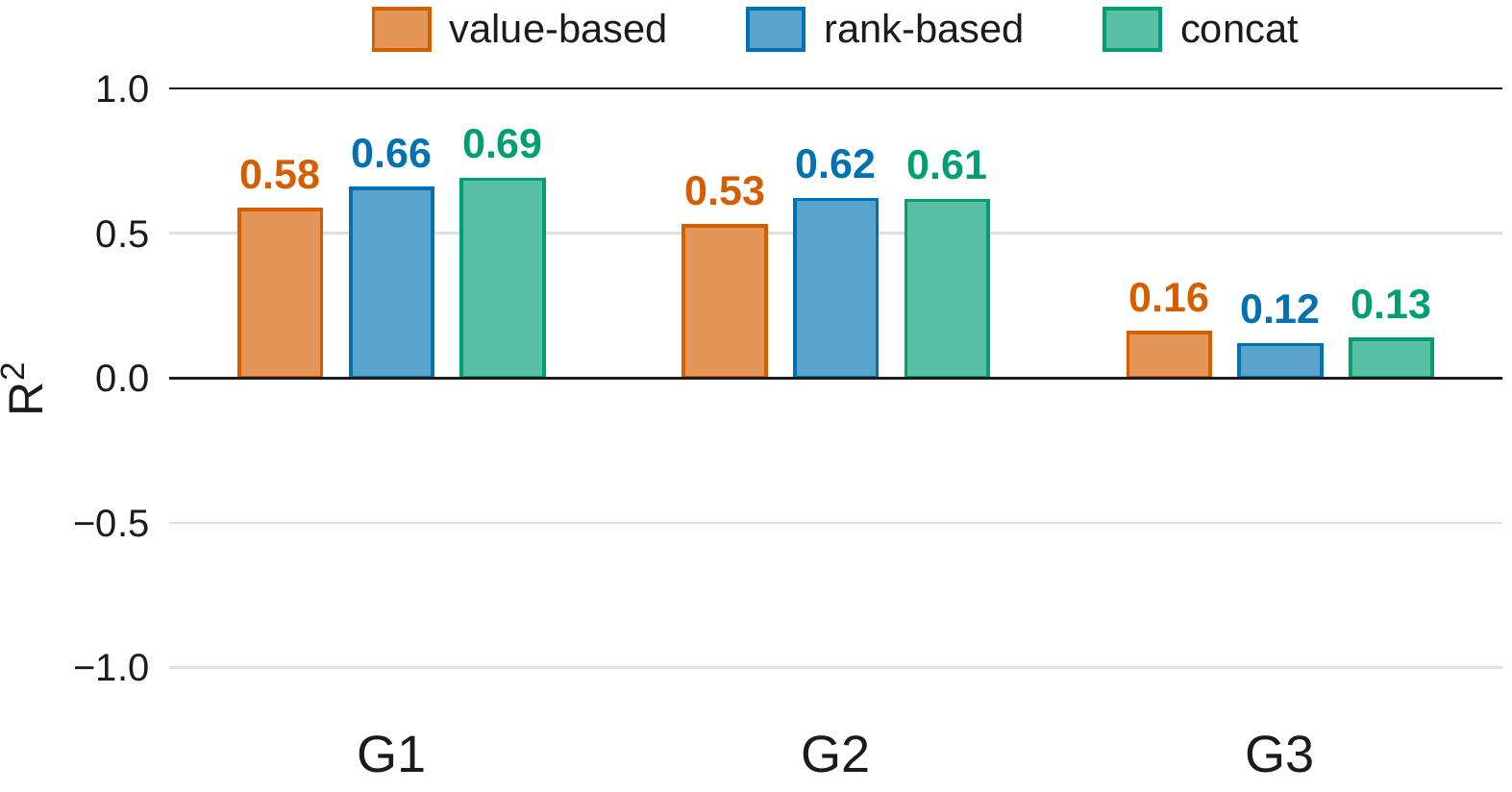}
    \caption{With data augmentation.}
    \label{fig:rq3-r2-ela-aug}
  \end{subfigure}
  \caption{Data augmentation raises the $R^{2}$ of value-based extraction and concatenation on the transformed test instances.}
  \Description{Two panels of R-squared values on transformed test instances for value-based, rank-based, and concatenated ELA feature extraction on the G1, G2, and G3 portfolios. Left: without data augmentation; right: with data augmentation. With augmentation, value-based extraction and concatenation rise from at most 0.28 to at least 0.53 for G1 and G2, and from negative to positive values for G3. Rank-based extraction barely changes for G1 and G2. After augmentation, concatenation is within 0.03 of rank-based extraction for G1 and G2.}
  \label{fig:rq3-r2}
\end{figure*}

Concatenation avoided this drawback but did not reduce the drop.
Its $R^{2}$ was within 0.01 of that of value-based extraction before the transformation and at most 0.25 for G1 and G2 after it.
Concatenation includes all rank-based features, but the learning model still relied on the information that the transformation changes.

In the selection task, rank-based extraction also gave a significantly higher $\Theta$ after the transformation (0.62, versus 0.60 for the others; \cref{fig:rq2-aas}), but the bootstrap interval of this advantage includes zero.
Before the transformation, rank-based extraction gave the lowest $\Theta$ (0.65, versus 0.70 for the others).

\subsection{RQ3: Does Data Augmentation Improve Prediction Performance on Transformed Instances?}
\label{sec:rq3}
Data augmentation significantly raised the $R^{2}$ of value-based extraction and concatenation on the transformed test instances for every group (\cref{fig:rq3-r2}).
For G1 and G2, it raised them from at most 0.28 to at least 0.53; for G3, from negative to positive values on average (\cref{tab:app-families}).
On the original test instances, it lowered these $R^{2}$ by 0.06--0.11 (\cref{tab:app-results}).
Doubling the training set with new untransformed instances did not give this gain (at most 0.27; Appendix~\ref{app:controls}).

After data augmentation, the $R^{2}$ of concatenation for G1 and G2 was higher than that of value-based extraction (\cref{fig:rq3-r2-ela-aug}).
Compared with rank-based extraction, it was significantly higher for G1 (0.69 vs.\ 0.66) but not for G2 (0.61 vs.\ 0.62); the bootstrap interval of the G1 difference includes zero.
Value-based extraction remained significantly below rank-based extraction (0.58 vs.\ 0.66 and 0.53 vs.\ 0.62).
A non-invariant feature can mix information that the algorithms use and ignore.
Data augmentation appears to have reduced the reliance on the latter (its predictions for G1 changed less; Appendix~\ref{app:controls}), so concatenation could match rank-based extraction.

For G1 and G2, data augmentation did not significantly change the $R^{2}$ of rank-based extraction (at most 0.01).
For G1, the added instances match the original ones in rank-based features and $J$.
New untransformed instances instead raised this $R^{2}$ to 0.74 and 0.71, above every extraction with data augmentation (Appendix~\ref{app:controls}).

For G3, data augmentation also raised the $R^{2}$ of rank-based extraction, from $-1.05$ to 0.12, but the drop remained.
The $R^{2}$ for G3 was at most 0.16 on the transformed test instances, and the $R^{2}$ of rank-based extraction on the original ones fell from 0.63 to 0.33 (\cref{tab:app-results}).
The gain of value-based extraction and concatenation for G3 depends on $\psi$ accompanying the rescaling, because only $\psi$ changes their input (Appendix~\ref{app:controls}).

\begin{figure}[t]
  \centering
  \includegraphics[width=\columnwidth]{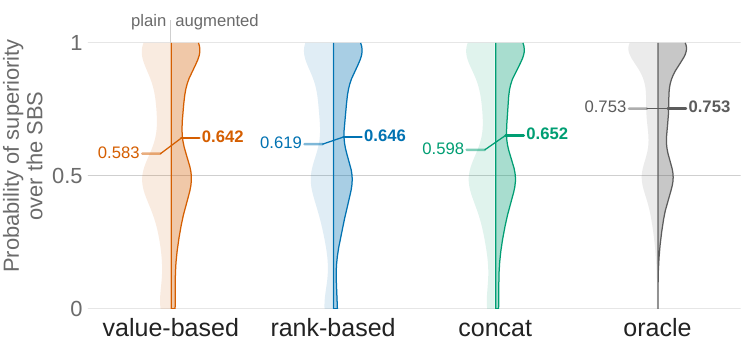}
  \caption{Data augmentation raises $\Theta$ on the transformed test instances for every feature extraction.
  In each violin, the right half is with data augmentation.}
  \Description{Split violin plots of the probability of superiority of AAS over the SBS on transformed test instances, for value-based, rank-based, and concatenated ELA feature extraction and the Oracle. Theta rises from 0.58 to 0.64, from 0.62 to 0.65, and from 0.60 to 0.65 for the three extractions. The Oracle has 0.75.}
  \label{fig:rq3-aas}
\end{figure}

In the selection task, data augmentation significantly raised $\Theta$ for every feature extraction, by 0.03--0.06 (\cref{fig:rq3-aas}).
After data augmentation, all three feature extractions reached 0.64--0.65; on the original test instances, data augmentation lowered the $\Theta$ of value-based extraction and concatenation by 0.02 (Appendix~\ref{app:controls}).

\section{Discussion}
\label{sec:discussion}
\begin{itemize}[leftmargin=*]
  \item \textbf{Diverse training instances may not be enough.}
  Even in diverse training instances, as in MA-BBOB~\cite{vermetten2025mabbob}, the labels can correlate with information that the algorithms ignore.
  Weakening this correlation may require instances that change such information but not $J$, as data augmentation does (\cref{sec:rq3}).
  \item \textbf{The test also applies to AAC and feature-free methods.}
  It requires only a model that predicts algorithm performance from objective values, whether the model also takes a configuration or takes the sampled solutions directly.
  \item \textbf{Limitation: shifts the remedies do not address.}
  The first kind keeps the order of objective values but changes algorithm performance, as the rescaling in \cref{eq:warp} does for G3.
  No remedy we studied removed the drop for G3 (\cref{sec:rq3}).
  The second kind changes the order itself, as on unseen functions, which the test does not cover.
  As in prior work on unseen problems~\cite{cenikj2025wall}, leave-one-function-out evaluation gave a median $R^{2}$ of at most $-0.16$ in every condition (Appendix~\ref{app:lofo}).
  \item \textbf{Limitation: scope of the evidence.}
  RQ1 is an existence result, for which one portfolio suffices.
  Its direct evidence is a G1 portfolio on 10-dimensional BBOB with one budget and sample size, LIO evaluation, the known minimum as $\hat{f}^{\min}_{p}$, the range $[0,1]$ in \cref{eq:warp}, and $\sigma = 1$.
  With functions also resampled, the RQ1 baseline comparison is inconclusive (Appendix~\ref{app:bootstrap}).
  \item \textbf{Future work.}
  Methods without supervised learning, such as those based on reinforcement learning~\cite{ma2023metabox} and large language models~\cite{yang2024large}, also receive objective values and should be tested.
\end{itemize}

\section{Conclusion}
\label{sec:conclusion}
We proposed the \testname{} and used it to show that shortcut learning can occur in feature-based AAS even with normalized objective values.
Because in-distribution accuracy can hide shortcut learning, we recommend reporting the test alongside it.
The \testname{} is available as the Python package \texttt{rubberruler}\footnote{\url{https://github.com/shuheitnk/rubberruler}}.
Rank-based feature extraction suits algorithms that search using ranks and requires no extra runs.
Data augmentation improved prediction on transformed instances for every portfolio we studied but requires extra algorithm runs; for G1 and G2, new instances with rank-based extraction gave a higher $R^{2}$ there (not tested for significance).
On the original instances, rank-based extraction lowered $R^{2}$ by 0.15--0.19, and data augmentation lowered it by up to 0.31.
Feature designers should ask which information the algorithms use and check the answer with the test.

\begin{acks}
JSPS KAKENHI Grant Number JP25K21301 supported this work.
Claude Code and Claude (Anthropic) assisted with implementing the experimental code and writing the paper.
The authors verified all code and content and take full responsibility for them.
\end{acks}

\bibliographystyle{ACM-Reference-Format}
\bibliography{references}

\clearpage
\onecolumn
\appendix
\section{Shortcut Learning Increases the Deployment Error}
\label{app:error}
This appendix shows that relying on the spurious feature (\cref{subsec:def-shortcut}) increases the deployment error.
Let $\bar{h}(\bm{z}_{\mathrm{core}}):=\mathbb{E}[h(\bm{z})\mid\bm{z}_{\mathrm{core}}]$ be the predictor averaged over the spurious feature, with all expectations under $\mathcal{D}_{\mathrm{deploy}}$.
Then
\begin{equation}
  \mathbb{E}\bigl[(h(\bm{z})-y)^{2}\bigr]
  =\mathbb{E}\bigl[(\bar{h}(\bm{z}_{\mathrm{core}})-y)^{2}\bigr]
  +\mathbb{E}\bigl[(h(\bm{z})-\bar{h}(\bm{z}_{\mathrm{core}}))^{2}\bigr],
  \label{eq:error-decomp}
\end{equation}
because the cross term vanishes by the definition of $\bar{h}$ and the second condition of \cref{eq:shortcut-def}.
The second term is the variation of the output caused by the spurious feature; it is positive whenever $h(\bm{z})
eq\bar{h}(\bm{z}_{\mathrm{core}})$ with positive probability, and we then say that $h$ relies on the spurious feature.
Hence a predictor that relies on the spurious feature has a larger deployment error than the same predictor with this reliance removed.
This argument requires the second condition of \cref{eq:shortcut-def} to hold exactly under $\mathcal{D}_{\mathrm{deploy}}$. The non-invariant features do not satisfy it as a whole, because they also carry information that stays valid under the transformation (\cref{sec:rq3}); the argument applies to the part of their information that loses its relation to $y$.
In the \testname{}, we take $\bm{z}_{\mathrm{core}}$ to be the information that the transformation does not change.
For G1, $J$ does not change either, so any change in the predictions comes from the reliance on $\bm{z}_{\mathrm{spurious}}$.

\FloatBarrier
\section{Examples of Transformations}
\label{app:warp}
\begin{figure}[!htbp]
  \centering
  \includegraphics[width=0.45\textwidth]{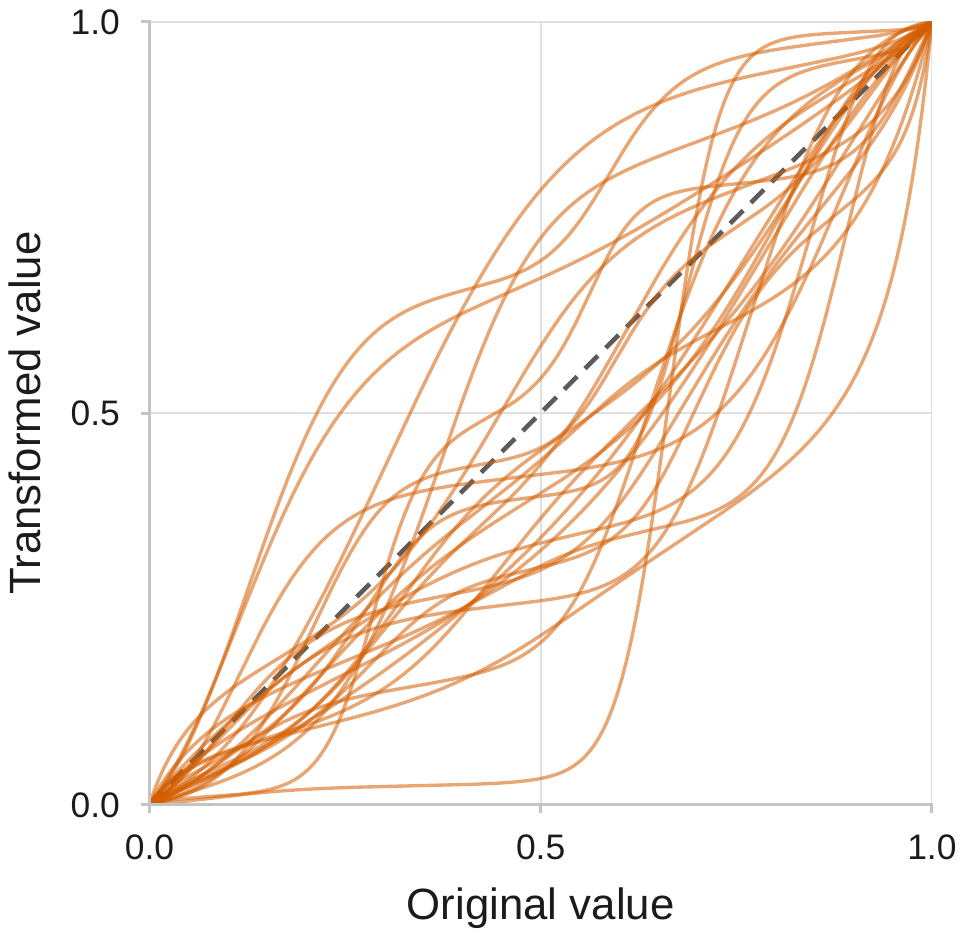}
  \caption{Functions $\psi$ drawn from $\Dspline$ warp normalized objective values nonlinearly but keep their order. The dashed line is the identity.}
  \Description{Increasing curves from (0,0) to (1,1) on the unit square, some bending above and some below the dashed identity line.}
  \label{fig:monotone-warp}
\end{figure}

\Cref{fig:monotone-warp} shows functions $\psi$ drawn from $\Dspline$ (\cref{subsec:transform}).

\FloatBarrier
\section{Experimental Details}
\label{app:settings}
All settings and instructions for reproduction are in our repository (\cref{subsec:overview}).

\paragraph{Algorithms and predictors.}
The restart rule based on the search scale replaces the default termination criteria of every algorithm, and we disable the criteria based on objective values, such as those of BIPOP CMA-ES.
SHADE and L-SHADE select solutions by comparison only but weight their parameter adaptation by the size of improvements in objective value.
BOBYQA fits a quadratic model to objective values, and Dual Annealing accepts solutions based on differences in objective value.
For the two-algorithm portfolios, $J(p,a_2)=1-J(p,a_1)$, so we train only the predictor for $a_1$ and predict $a_2$ by $1-\hat{J}(p,a_1)$; for the mixed portfolio, we train one predictor per algorithm.
On the sampled solutions, rank-based input equals a strictly increasing transformation of the objective values, so it is a valid input to Deep-ELA.

\paragraph{Distributions in RQ3.}
The four distributions that data augmentation does not use are $\Dspline$ with $K\in\{2,3\}$, $\Dspline$ with $K\in\{14,\dots,19\}$, power functions $\psi(t)=t^{\gamma}$, and piecewise-linear functions.
For power functions, $\log\gamma\sim\mathcal{N}(0,1)$.
For piecewise-linear functions, we draw the number of interior breakpoints uniformly from $\{4,\dots,7\}$, their positions uniformly from $(0,1)$, and the slope of each interval from $\exp\bigl(\mathcal{N}(0,1)\bigr)$, and normalize so that $\psi(1)=1$.

\paragraph{Families for Holm correction.}
We correct ELA and Deep-ELA separately.
In RQ2, a family is the three pairwise comparisons of feature extractions for each portfolio.
In RQ3, a family is the three pairwise comparisons for each portfolio and training condition; for data augmentation, a family is the nine cells (three portfolios $\times$ three extractions) in the regression task and the three extractions in the selection task.

\begin{figure}[!htbp]
  \centering
  \includegraphics[width=0.8\textwidth]{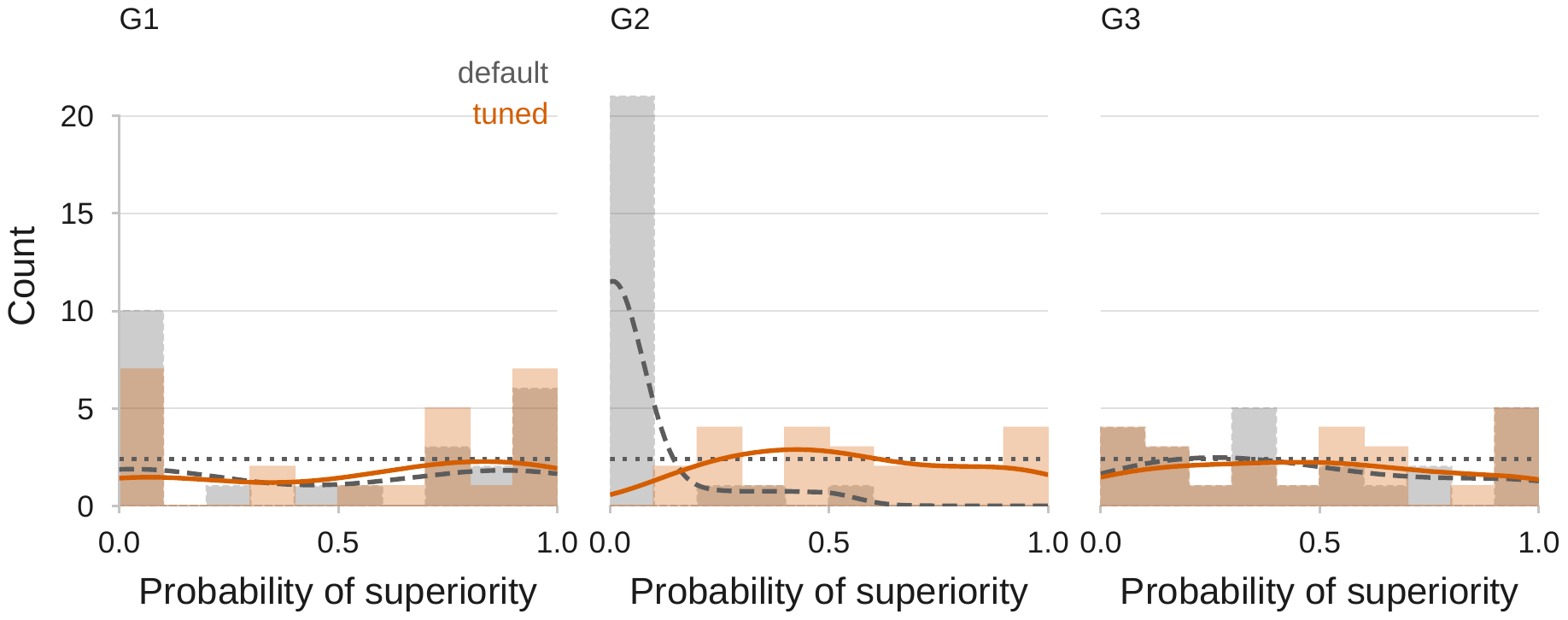}
  \caption{Tuning brought the distribution of the probability of superiority closer to uniform. Histograms before and after tuning over the 24 functions used for tuning in the regression task; the horizontal dashed line is the uniform distribution.}
  \Description{Histograms of the probability of superiority before and after tuning for the three algorithm pairs; after tuning the bars are closer to the uniform level.}
  \label{fig:app-tuning}
\end{figure}

\paragraph{Tuned parameters.}
\Cref{tab:app-pair-selected,tab:app-role-selected} list the tuned parameters; all others are published settings or library defaults.
All tuning used instance 11 of each of the 24 functions, which is a training instance, with 31 runs per configuration.
In the regression task, we tuned only the weaker algorithm of each pair.
We chose the parameters that minimize the Kolmogorov--Smirnov distance between the distribution of the probability of superiority over the 24 functions and the uniform distribution.
The purpose is to spread the regression target over $[0,1]$ (\cref{fig:app-tuning}).
In the selection task, we tuned all six algorithms over one grid of 1,230 configurations.
The score of a configuration on a function is its mean probability of superiority over the other configurations in the grid.
For selection (budget $1000d$), we chose the combination of configurations that maximizes the mean over the 24 functions of the highest score among the six algorithms.
For the SBS (budget $1050d$), we chose for each algorithm the configuration with the highest mean score.
The SBS is then the algorithm with the highest mean $J$ over the training instances, which is BIPOP CMA-ES.

\begin{table}[!htbp]
  \centering
  \caption{Tuned algorithms and parameters in the regression task. The offset is the shift of the initial simplex vertices from the start point relative to the width of the search range.}
  \label{tab:app-pair-selected}
  \small
  \begin{tabular}{@{}ll@{}}
    \toprule
    Algorithm & Selected parameters \\
    \midrule
    Adaptive Nelder--Mead & offset 0.25, $\varepsilon_{x} = 10^{-4}$ \\
    SHADE & $N = 5d$, $M_{F} = 0.7$, $M_{CR} = 0.4$, $\varepsilon_{x} = 10^{-8}$ \\
    Dual Annealing & $T_{0} = 5230$, $q_{v} = 1.5$, $q_{a} = -5$, $\varepsilon_{x} = 10^{-8}$ \\
    \bottomrule
  \end{tabular}
\end{table}

\begin{table}[!htbp]
  \centering
  \caption{Parameters selected for the selection task. The SBS uses BIPOP CMA-ES tuned separately for it. The initial step of BOBYQA is relative to the width of the search range.}
  \label{tab:app-role-selected}
  \small
  \begin{tabular}{@{}lll@{}}
    \toprule
    Algorithm & For selection (budget $1000d$) & For the SBS (budget $1050d$) \\
    \midrule
    BIPOP CMA-ES & $\sigma_{0} = 4.0$, $\lambda_{0} = 30$, $\varepsilon_{x} = 10^{-8}$ & $\sigma_{0} = 5.0$, $\lambda_{0} = 30$, $\varepsilon_{x} = 10^{-8}$ \\
    Adaptive Nelder--Mead & offset 0.25, $\varepsilon_{x} = 10^{-8}$ & offset 0.25, $\varepsilon_{x} = 10^{-8}$ \\
    SHADE & $N = 5d$, $M_{F} = 0.5$, $M_{CR} = 0.4$, $\varepsilon_{x} = 10^{-8}$ & $N = 5d$, $M_{F} = 0.4$, $M_{CR} = 0.7$, $\varepsilon_{x} = 10^{-8}$ \\
    L-SHADE & $N_{\mathrm{init}} = 5d$, $M_{F} = 0.3$, $M_{CR} = 0.4$, $\varepsilon_{x} = 10^{-8}$ & $N_{\mathrm{init}} = 5d$, $M_{F} = 0.3$, $M_{CR} = 0.6$, $\varepsilon_{x} = 10^{-8}$ \\
    BOBYQA & initial step 0.4, $\varepsilon_{x} = 10^{-4}$ & initial step 0.4, $\varepsilon_{x} = 10^{-6}$ \\
    Dual Annealing & $T_{0} = 40000$, $q_{v} = 2.85$, $q_{a} = -5$, $\varepsilon_{x} = 10^{-6}$ & $T_{0} = 500$, $q_{v} = 1.5$, $q_{a} = -50$, $\varepsilon_{x} = 10^{-6}$ \\
    \bottomrule
  \end{tabular}
\end{table}

\paragraph{Invariant features.}
\Cref{tab:app-ela-groups} lists the invariant features of each group, chosen from the feature definitions.
In the preliminary analysis (\cref{sec:pre_study}), these 36 features did not change and the other 23 did.
We removed two features because they take a non-finite value on some instance: \texttt{ela\_meta.\allowbreak quad\_simple.\allowbreak cond} on $f_5$, and \texttt{ic.eps\_ratio} on test instances transformed by power functions.
This follows the rule in \cref{subsec:exp-features}, which uses the features of the test instances but not their $J$.
We apply the rule over all training and test instances so that every experiment uses the same 59 features; over the training instances alone, it would remove only the first of the two.

\begin{table}[!htbp]
  \centering
  \caption{Number of all and invariant ELA features in each group.}
  \label{tab:app-ela-groups}
  \small
  \begin{tabular}{@{}lrl@{}}
    \toprule
    Group & No. & Invariant features \\
    \midrule
    \texttt{disp} & 16 & all (16) \\
    \texttt{ela\_level} & 9 & all (9) \\
    \texttt{ela\_meta} & 8 & none \\
    \texttt{pca} & 8 & the 4 computed from coordinates only (ending in \texttt{\_x}) \\
    \texttt{fitness\_distance} & 6 & \texttt{distance\_mean}, \texttt{distance\_std} (2) \\
    \texttt{nbc} & 5 & all but \texttt{nb\_fitness.cor} (4) \\
    \texttt{ic} & 4 & \texttt{m0} (1) \\
    \texttt{ela\_distr} & 3 & none \\
    \midrule
    Total & 59 & 36 \\
    \bottomrule
  \end{tabular}
\end{table}

\FloatBarrier
\section{Preliminary Analysis}
\label{app:pre-study}
This appendix checks two preconditions for non-invariant features to act as spurious features: strictly increasing transformations change these features, but they do not change $J$ of G1.
We compare each test instance $p\in\Ptest$ with its transformed instance $\tilde{p} := g_{\psi_{p},p}\circ p$ (\cref{subsec:test-procedure}).
For each of G1, G2, and G3, we compute $J$ of the two algorithms in the group.
We measure the change $\Delta_{\eta}$ of a feature or of $J$ with \cref{eq:variation} of the main paper.
Dividing by the standard deviation lets us compare features and $J$ on the same scale.
For features, we report the median of $\Delta_{\eta}$ over the 11 seeds $\omega$.
\Cref{sec:pre_study} of the main paper reports the results (\cref{fig:pre-variation}).
The change in $J$ increases from G1 to G3.
This order matches how the algorithms use objective values: G2 uses them only to adapt its parameters, whereas G3 uses them directly in its search.
The rescaling in \cref{eq:warp} alone changes $J$ of G3 by 0.65 (Appendix~\ref{app:controls}).

\FloatBarrier
\section{Deep-ELA Results and All Numbers}
\label{app:deep-ela}
\begin{figure}[!htbp]
  \centering
  \includegraphics[width=0.8\textwidth]{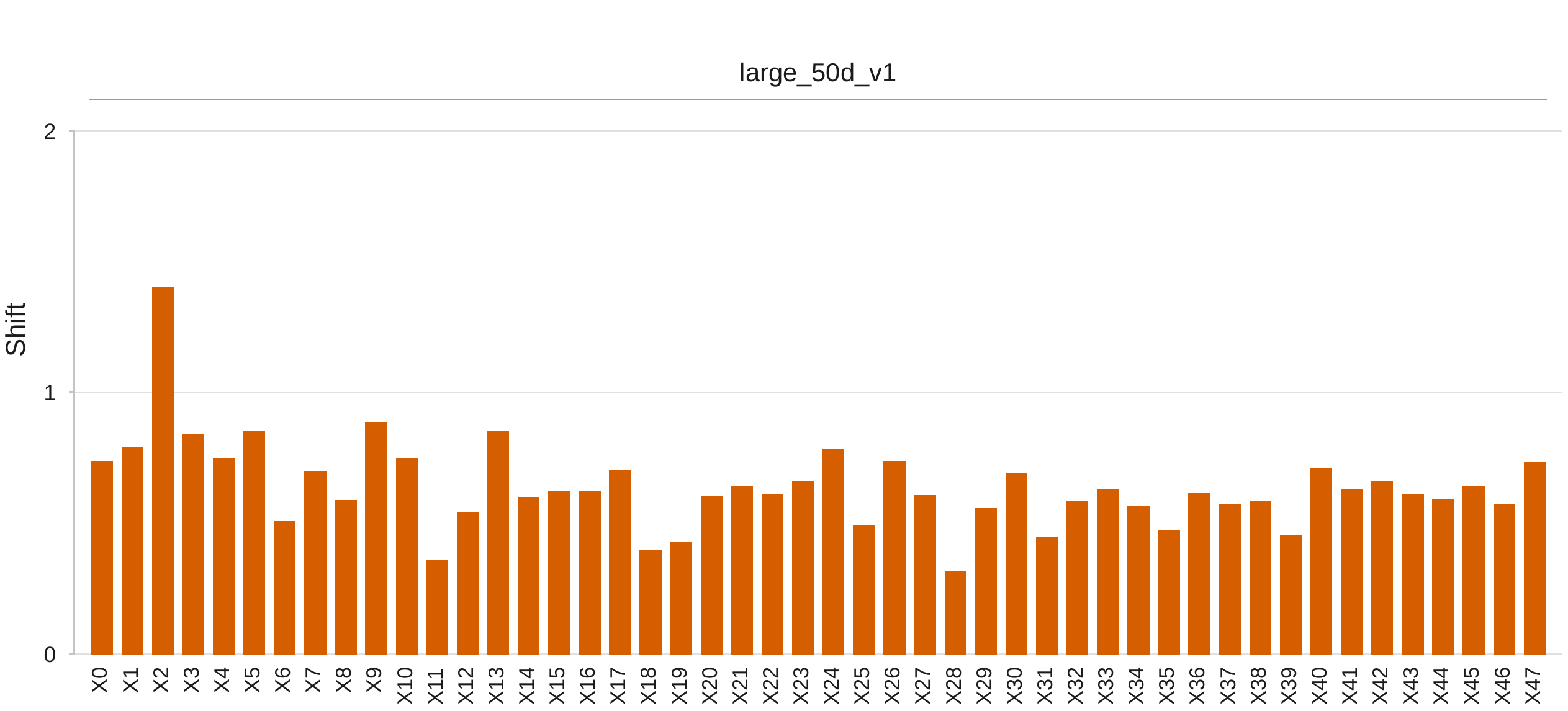}
  \caption{Strictly increasing transformations change all 48 Deep-ELA features. Axes as in \cref{fig:pre-variation}.}
  \Description{Bar chart of the shift of the 48 Deep-ELA features under the transformation; all are clearly above zero, with a median of 0.62.}
  \label{fig:app-deep-ela-shift}
\end{figure}

This appendix repeats RQ2 and RQ3 of the main paper with the 48 features of the pretrained Deep-ELA instead of the 59 ELA features.
We use Deep-ELA without retraining it, and all other settings are the same as in the main paper.
We do not use Deep-ELA in RQ1, because it has no invariant features: the transformation changed all 48 of its features, with a median $\Delta_{\eta}$ of 0.62 (\cref{fig:app-deep-ela-shift}).

\begin{figure}[!htbp]
  \centering
  \begin{subfigure}[t]{0.48\textwidth}
    \centering
    \includegraphics[width=\linewidth]{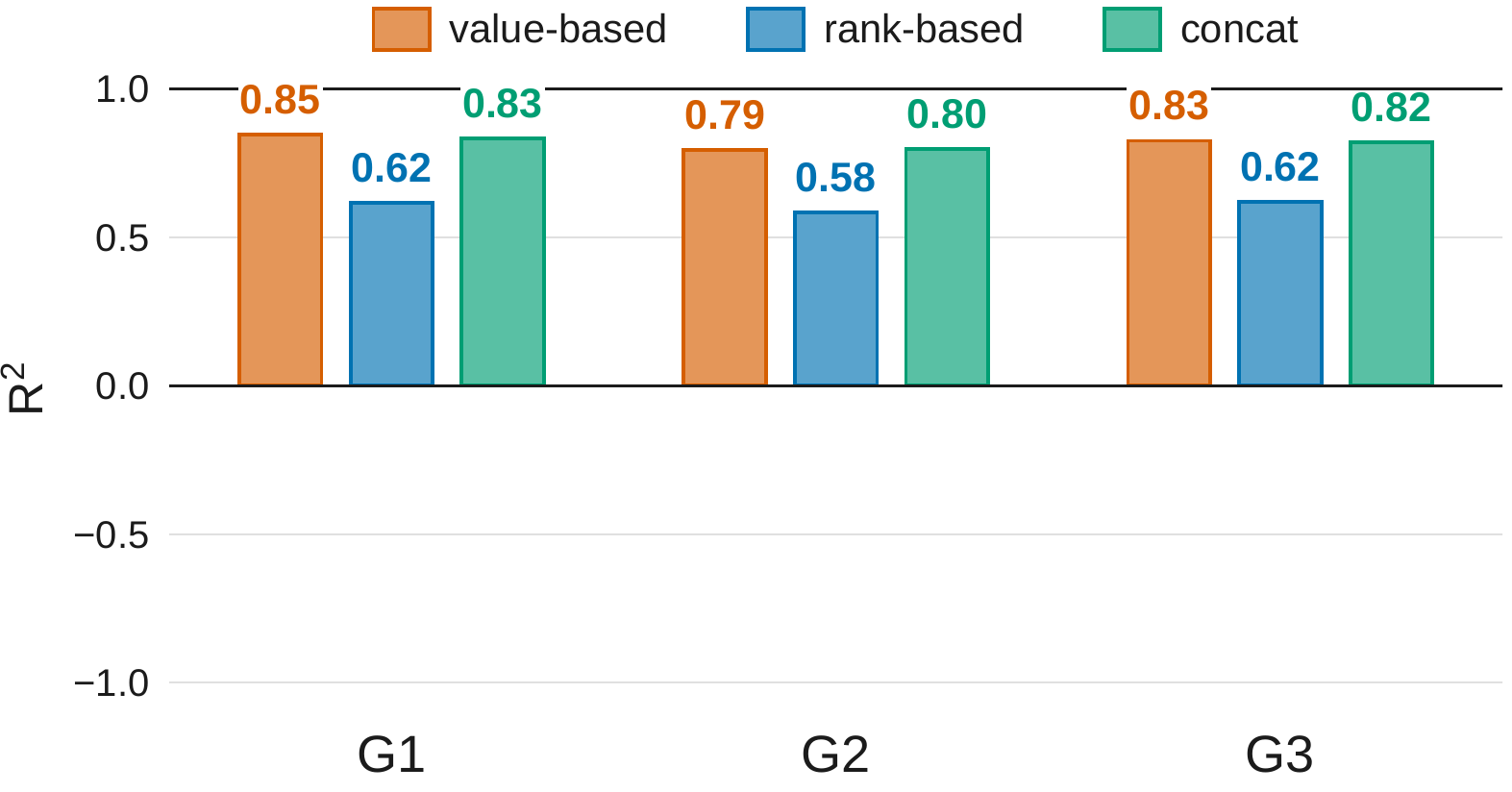}
    \caption{Original test instances.}
  \end{subfigure}\hfill
  \begin{subfigure}[t]{0.48\textwidth}
    \centering
    \includegraphics[width=\linewidth]{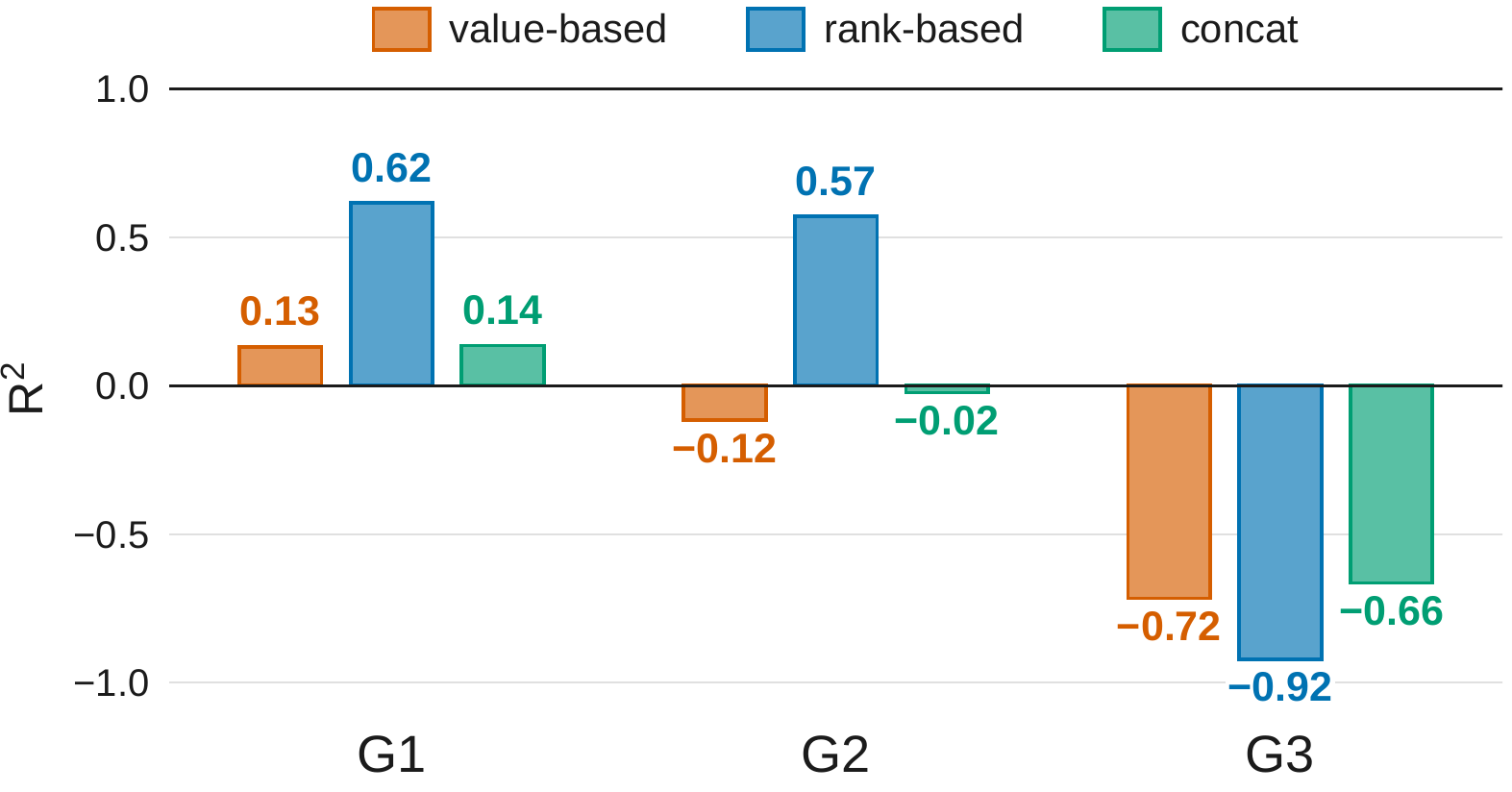}
    \caption{Transformed test instances.}
  \end{subfigure}
  \caption{RQ2 with Deep-ELA. Rank-based feature extraction gives the lowest $R^{2}$ on the original test instances but the highest $R^{2}$ for G1 and G2 on the transformed test instances.}
  \Description{Two panels of R-squared values for value-based, rank-based, and concatenated Deep-ELA feature extraction on the G1, G2, and G3 portfolios. Left: original test instances; right: transformed test instances. On the original instances, rank-based extraction is the lowest in every group. On the transformed instances, rank-based extraction stays at 0.57 or higher for G1 and G2, whereas the other two drop to 0.14 or lower; all three are negative for G3.}
  \label{fig:app-deep-ela-rq2}
\end{figure}

\begin{figure}[!htbp]
  \centering
  \begin{subfigure}[t]{0.48\textwidth}
    \centering
    \includegraphics[width=\linewidth]{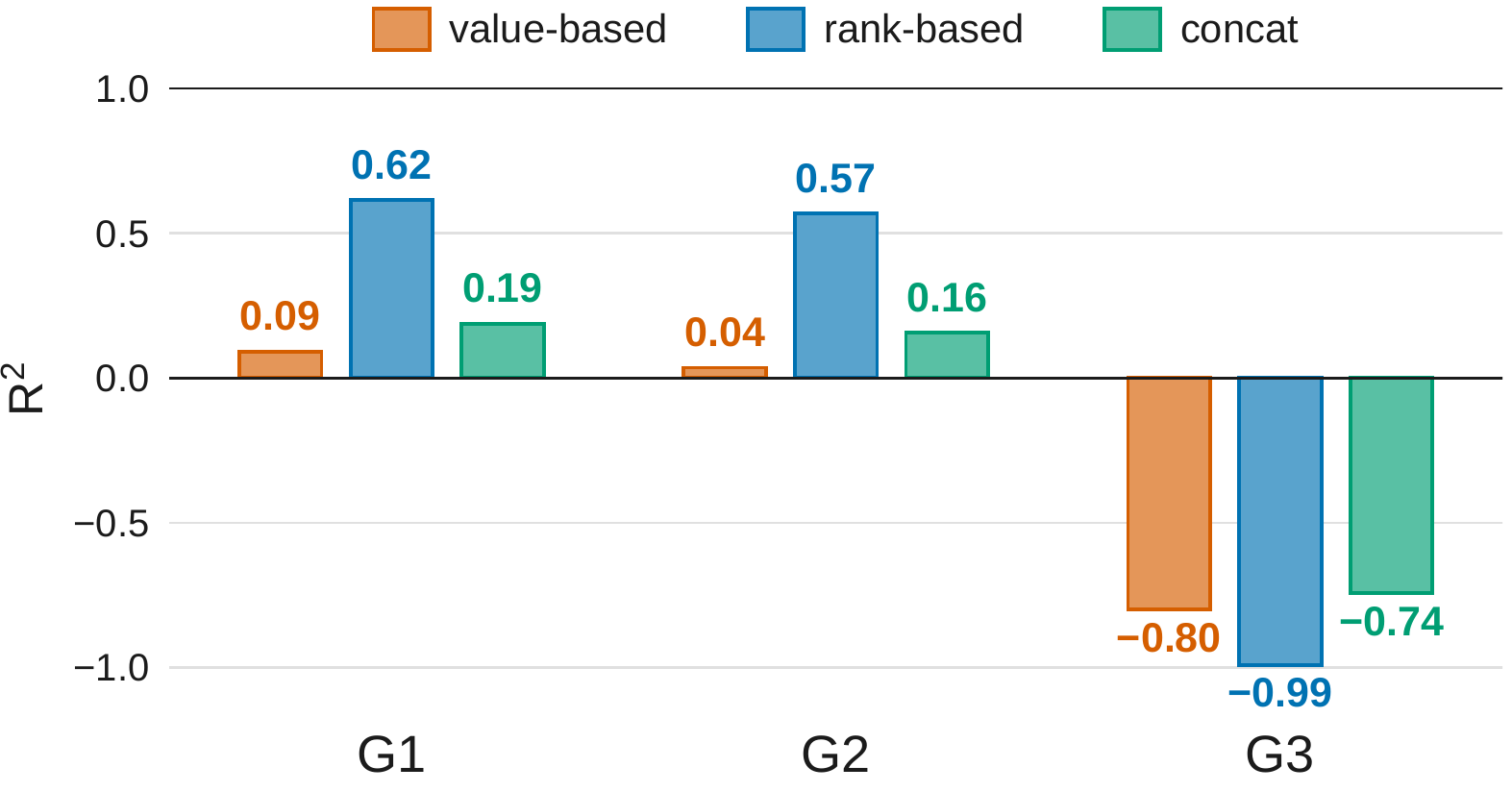}
    \caption{Without data augmentation.}
  \end{subfigure}\hfill
  \begin{subfigure}[t]{0.48\textwidth}
    \centering
    \includegraphics[width=\linewidth]{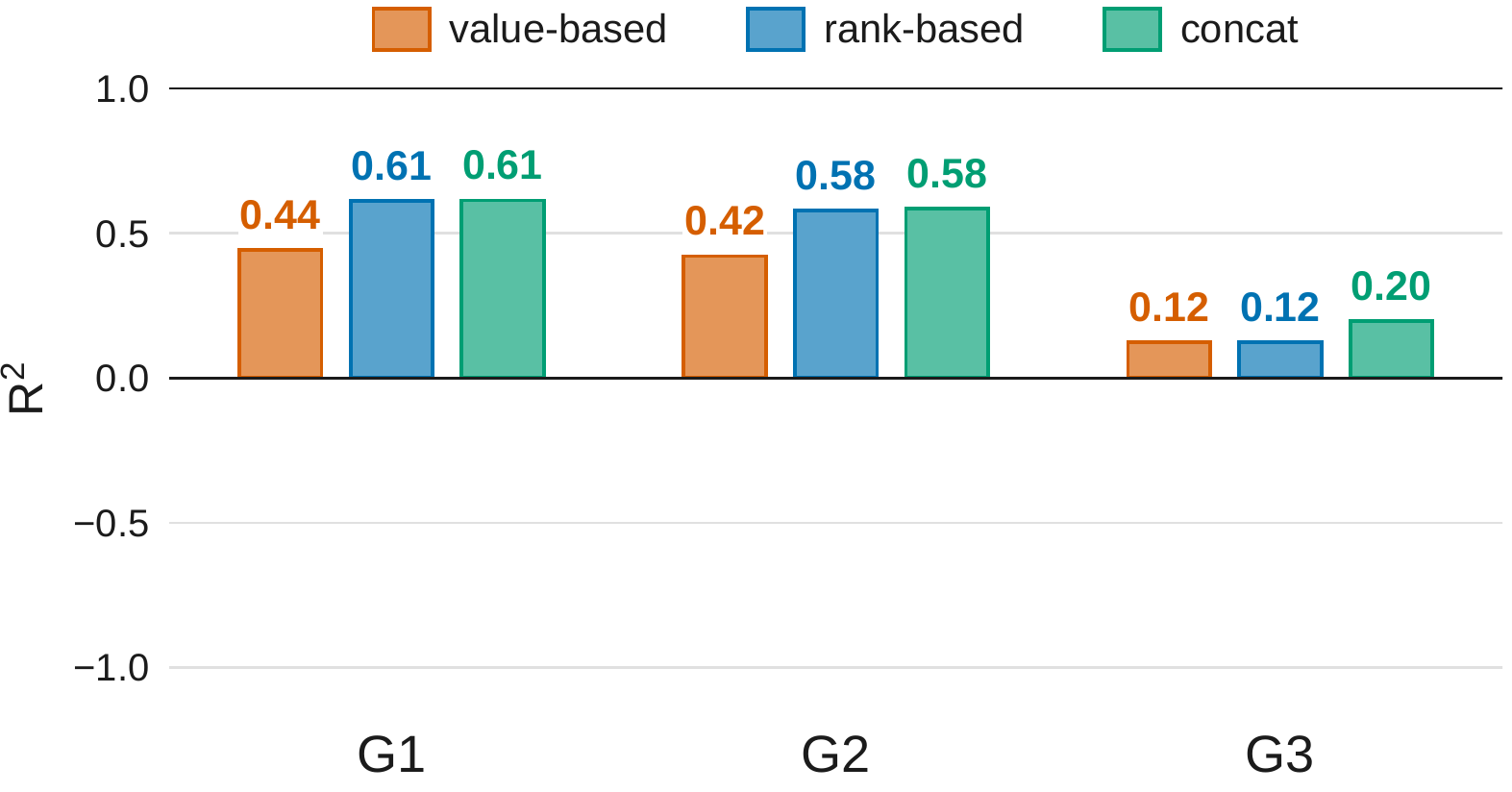}
    \caption{With data augmentation.}
  \end{subfigure}
  \caption{RQ3 with Deep-ELA. Data augmentation raises the $R^{2}$ of value-based extraction and concatenation on the transformed test instances. The test instances are transformed with four distributions that data augmentation does not use.}
  \Description{Two panels of R-squared values on transformed test instances for value-based, rank-based, and concatenated Deep-ELA feature extraction on the G1, G2, and G3 portfolios. Left: without data augmentation; right: with data augmentation. With augmentation, value-based extraction and concatenation rise for every group, and from negative to positive values for G3. Rank-based extraction barely changes for G1 and G2.}
  \label{fig:app-deep-ela-rq3}
\end{figure}

\begin{figure}[!htbp]
  \centering
  \begin{subfigure}[t]{0.48\textwidth}
    \centering
    \includegraphics[width=\linewidth]{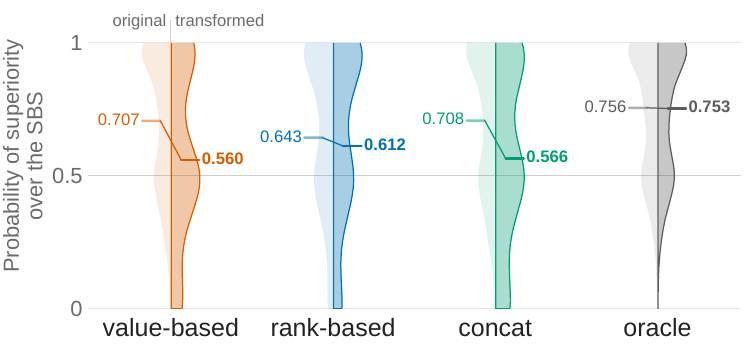}
    \caption{RQ2. Left halves: original; right halves: transformed.}
  \end{subfigure}\hfill
  \begin{subfigure}[t]{0.48\textwidth}
    \centering
    \includegraphics[width=\linewidth]{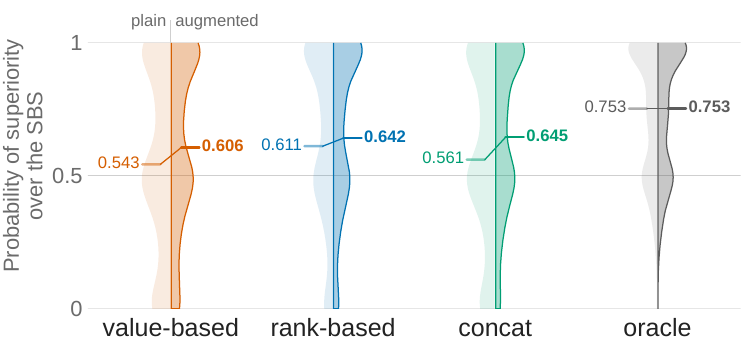}
    \caption{RQ3. Left halves: without data augmentation; right halves: with data augmentation (transformed test instances).}
  \end{subfigure}
  \caption{Selection task with Deep-ELA. On the transformed test instances, rank-based feature extraction gives the highest $\Theta$ (left), and data augmentation raises $\Theta$ for every feature extraction (right). The numbers are $\Theta$.}
  \Description{Two panels of split violin plots of the probability of superiority of AAS over the SBS for value-based, rank-based, and concatenated Deep-ELA feature extraction and the Oracle. Left: original to transformed, value 0.71 to 0.56, rank 0.64 to 0.61, concat 0.71 to 0.57. Right: without to with augmentation, value 0.54 to 0.61, rank 0.61 to 0.64, concat 0.56 to 0.65.}
  \label{fig:app-deep-ela-aas}
\end{figure}

Deep-ELA led to the same conclusions as ELA in RQ2 and RQ3.
In RQ2, rank-based feature extraction gave the lowest $R^{2}$ on the original test instances but the highest $R^{2}$ for G1 and G2 on the transformed test instances (\cref{fig:app-deep-ela-rq2}).
In the selection task, it also gave the highest $\Theta$ on the transformed test instances, with a larger advantage than with ELA (\cref{fig:app-deep-ela-aas}).
In RQ3, data augmentation raised the $R^{2}$ of value-based extraction and concatenation on the transformed test instances (\cref{fig:app-deep-ela-rq3}) and raised $\Theta$ for every feature extraction (\cref{fig:app-deep-ela-aas}).

\paragraph{All numbers.}
\Cref{tab:app-results} lists all results of RQ2 and RQ3 for ELA and Deep-ELA.

\begin{table}[!htbp]
  \centering
  \caption{All results of RQ2 and RQ3.
    $R^2$ (G1--G3, regression) and $\Theta$ (mixed portfolio, selection) are medians over 11 model seeds.
    \emph{Original}: original test instances. \emph{Transformed}: test instances transformed by $\Dspline$ (RQ2).
    \emph{RQ3}: test instances transformed by the four distributions that data augmentation does not use, without (\emph{plain}) and with (\emph{aug.}) data augmentation; \emph{Original, aug.}: original test instances with data augmentation ($\Theta$: \cref{tab:app-doubled}).
    Bold marks the best feature extraction in each column of each block.
    The Oracle reaches $\Theta=0.756$ on the original and $0.753$ on the transformed test instances.}
  \label{tab:app-results}
  \small
  \setlength{\tabcolsep}{4.5pt}
  \begin{tabular}{@{}llrrrrrrrr@{}}
    \toprule
    & & \multicolumn{4}{c}{ELA} & \multicolumn{4}{c}{Deep-ELA} \\
    \cmidrule(lr){3-6}\cmidrule(lr){7-10}
    Test instances & Extraction & G1 & G2 & G3 & $\Theta$ & G1 & G2 & G3 & $\Theta$ \\
    \midrule
    Original & value  & \textbf{0.85} & \textbf{0.77} & \textbf{0.78} & \textbf{0.704} & \textbf{0.85} & 0.79 & \textbf{0.83} & 0.707 \\
             & rank   & 0.66 & 0.62 & 0.63 & 0.646 & 0.62 & 0.58 & 0.62 & 0.643 \\
             & concat & \textbf{0.85} & 0.76 & \textbf{0.78} & \textbf{0.704} & 0.83 & \textbf{0.80} & 0.82 & \textbf{0.708} \\
    \midrule
    Transformed & value  & 0.23 & 0.20 & \textbf{$-$0.80} & 0.598 & 0.13 & $-$0.12 & $-$0.72 & 0.560 \\
                & rank   & \textbf{0.66} & \textbf{0.61} & $-$0.95 & \textbf{0.620} & \textbf{0.62} & \textbf{0.57} & $-$0.92 & \textbf{0.612} \\
                & concat & 0.25 & 0.18 & \textbf{$-$0.80} & 0.600 & 0.14 & $-$0.02 & \textbf{$-$0.66} & 0.566 \\
    \midrule
    RQ3, plain & value  & 0.24 & 0.25 & \textbf{$-$0.81} & 0.583 & 0.09 & 0.04 & $-$0.80 & 0.543 \\
               & rank   & \textbf{0.66} & \textbf{0.61} & $-$1.05 & \textbf{0.619} & \textbf{0.62} & \textbf{0.57} & $-$0.99 & \textbf{0.611} \\
               & concat & 0.26 & 0.28 & $-$0.83 & 0.598 & 0.19 & 0.16 & \textbf{$-$0.74} & 0.561 \\
    \midrule
    RQ3, aug.  & value  & 0.58 & 0.53 & \textbf{0.16} & 0.642 & 0.44 & 0.42 & 0.12 & 0.606 \\
               & rank   & 0.66 & \textbf{0.62} & 0.12 & 0.646 & \textbf{0.61} & \textbf{0.58} & 0.12 & 0.642 \\
               & concat & \textbf{0.69} & 0.61 & 0.13 & \textbf{0.652} & \textbf{0.61} & \textbf{0.58} & \textbf{0.20} & \textbf{0.645} \\
    \midrule
    Original, aug. & value  & 0.75 & 0.68 & 0.71 & --- & \textbf{0.82} & \textbf{0.76} & \textbf{0.73} & --- \\
                   & rank   & 0.66 & 0.63 & 0.33 & --- & 0.61 & 0.59 & 0.33 & --- \\
                   & concat & \textbf{0.76} & \textbf{0.70} & \textbf{0.72} & --- & 0.79 & 0.74 & 0.70 & --- \\
    \bottomrule
  \end{tabular}
\end{table}

\FloatBarrier
\section{Other Learning Models}
\label{app:models}
We replaced the learning model with random forest (scikit-learn defaults~\cite{pedregosa2011scikit}) and XGBoost (the tuned defaults XGB-TD of Holzmüller et al.~\cite{holzmuller2024better}); all other settings are the same as in the main paper.
With both models, the main conclusions of RQ1--RQ3 did not change: shortcut learning in RQ1, the advantage of rank-based extraction for G1 and G2 in RQ2, and the gain from data augmentation for value-based extraction and concatenation in RQ3.
Some other comparisons differed in significance.
Most of them were not significant with RealMLP, such as concatenation versus value-based extraction in the selection task; with XGBoost and ELA, rank-based extraction for G3 was no longer significantly below value-based extraction.
\Cref{tab:app-models} lists the levels.

\begin{figure}[!htbp]
  \centering
  \begin{subfigure}[t]{0.4\textwidth}
    \centering
    \includegraphics[width=\linewidth]{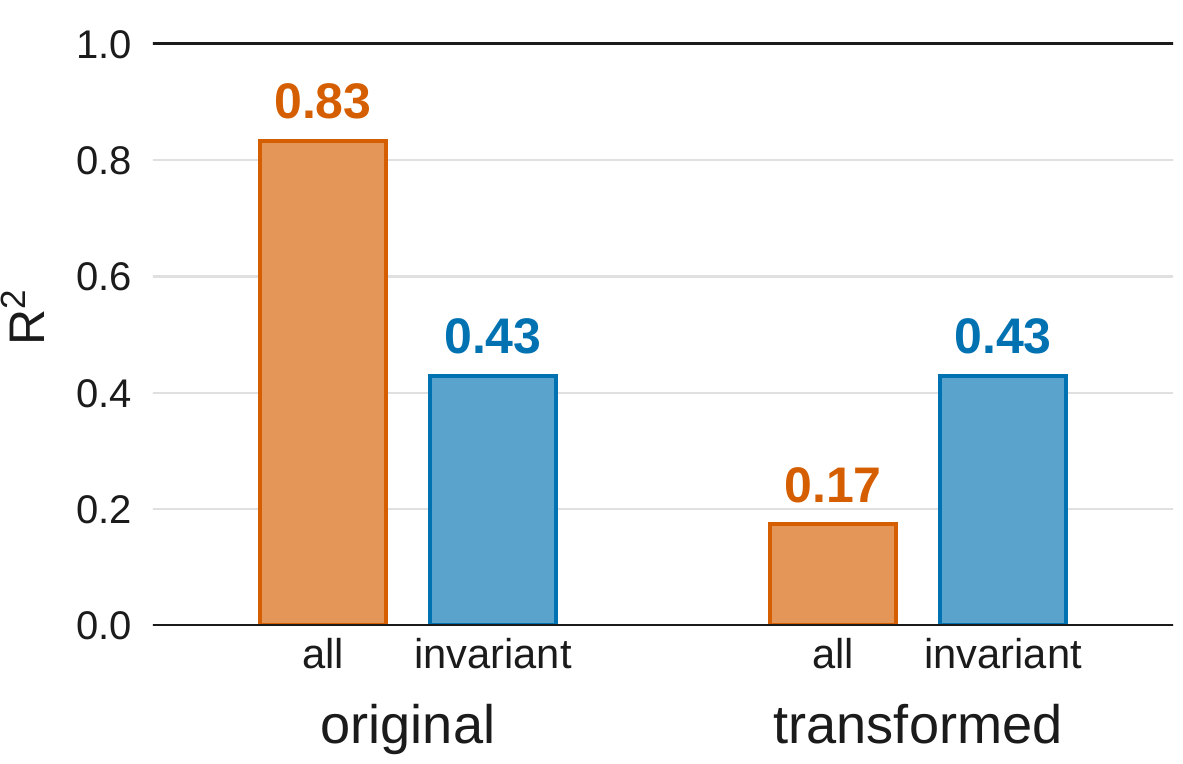}
    \caption{Random forest.}
  \end{subfigure}\hfill
  \begin{subfigure}[t]{0.4\textwidth}
    \centering
    \includegraphics[width=\linewidth]{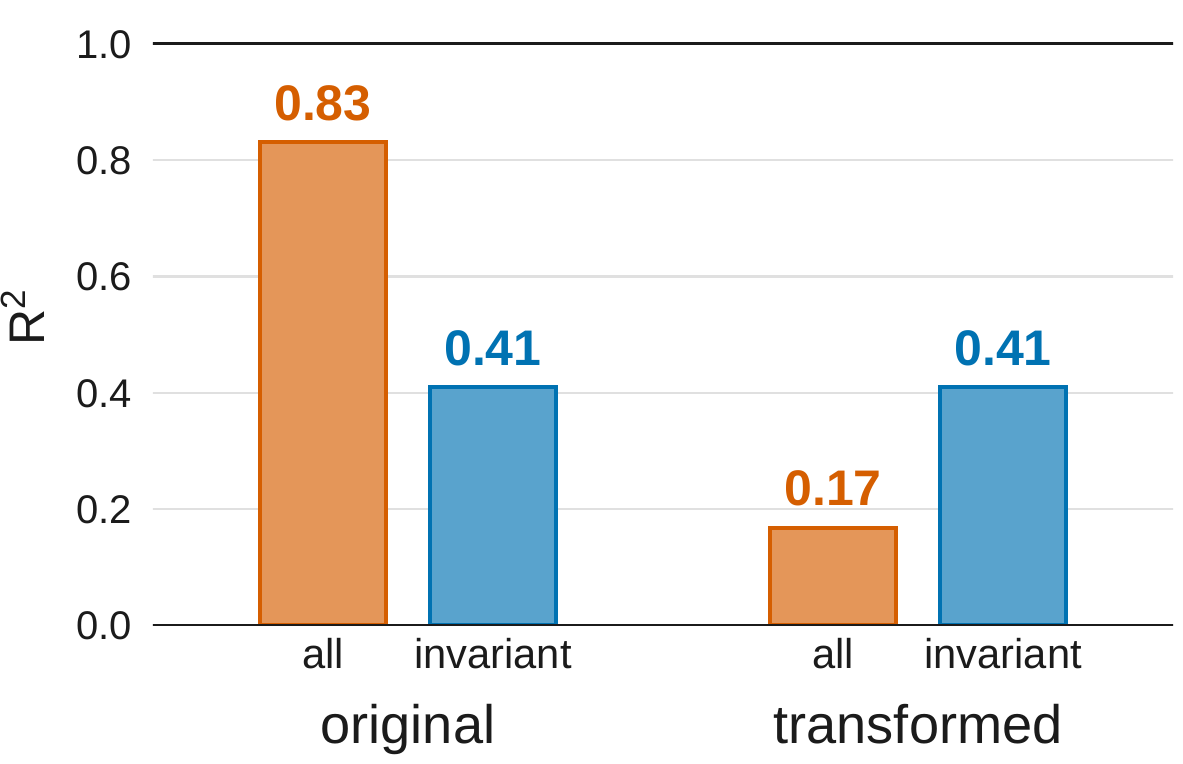}
    \caption{XGBoost.}
  \end{subfigure}
  \caption{Shortcut learning in the G1 regression task also occurs with random forest and XGBoost. Axes as in the right panel of \cref{fig:teaser}.}
  \Description{Two bar charts like the right part of Figure 1 of the paper: all features drop from 0.83 to 0.17 after the transformation, while invariant features stay at 0.43 (random forest) and 0.41 (XGBoost).}
  \label{fig:app-models-rq1}
\end{figure}

\paragraph{Random forest.}
\begin{itemize}
  \item RQ1: the $R^2$ of the predictor with all features dropped from 0.83 to 0.17, whereas that of the baseline stayed at 0.43 (\cref{fig:app-models-rq1}).
  \item RQ2: on the transformed test instances, the $R^2$ of rank-based extraction for G1 and G2 was at least 0.57, versus at most 0.25 for value-based extraction and concatenation; for G3, all extractions had a negative $R^2$. Before the transformation, the $R^2$ of rank-based extraction was lower than that of value-based extraction.
  \item RQ3: data augmentation reduced the drop of value-based extraction and concatenation for every group (from negative to positive values for G3); value-based extraction stayed below rank-based extraction for G1 and G2; data augmentation raised $\Theta$ for every feature extraction.
\end{itemize}

\paragraph{XGBoost.}
\begin{itemize}
  \item RQ1: the $R^2$ of the predictor with all features dropped from 0.83 to 0.17, whereas that of the baseline stayed at 0.41 (\cref{fig:app-models-rq1}).
  \item RQ2: on the transformed test instances, the $R^2$ of rank-based extraction for G1 and G2 was at least 0.57, versus at most 0.22 for value-based extraction and concatenation; for G3, all extractions had a negative $R^2$. Before the transformation, the $R^2$ of rank-based extraction was lower than that of value-based extraction.
  \item RQ3: the same as for random forest.
\end{itemize}

\begin{table}[!htbp]
  \centering
  \caption{Levels with random forest and XGBoost (medians over 11 model seeds). RQ1: $R^{2}$ for G1 on the original and the transformed test instances. RQ2: $R^{2}$ and $\Theta$ on the test instances transformed by $\Dspline$. RQ3: $R^{2}$ and $\Theta$ on the test instances transformed by the four distributions that data augmentation does not use, without / with data augmentation.}
  \label{tab:app-models}
  \footnotesize
  \begin{tabular}{@{}lllcccc@{}}
    \toprule
    & & & \multicolumn{2}{c}{Random forest} & \multicolumn{2}{c}{XGBoost} \\
    \cmidrule(lr){4-5}\cmidrule(lr){6-7}
    & Task & Extraction & ELA & Deep-ELA & ELA & Deep-ELA \\
    \midrule
    RQ1 & G1 & all features, original & 0.834 & -- & 0.831 & -- \\
     & G1 & all features, transformed & 0.174 & -- & 0.167 & -- \\
     & G1 & invariant features & 0.429 & -- & 0.410 & -- \\
    \midrule
    RQ2 & G1 & value & 0.174 & 0.195 & 0.167 & 0.176 \\
     &  & rank & 0.621 & 0.575 & 0.612 & 0.569 \\
     &  & concat & 0.160 & 0.252 & 0.179 & 0.219 \\
     & G2 & value & 0.099 & 0.026 & 0.082 & $-0.020$ \\
     &  & rank & 0.567 & 0.578 & 0.571 & 0.570 \\
     &  & concat & 0.093 & 0.084 & 0.091 & 0.011 \\
     & G3 & value & $-0.735$ & $-0.609$ & $-0.756$ & $-0.664$ \\
     &  & rank & $-0.767$ & $-0.790$ & $-0.794$ & $-0.821$ \\
     &  & concat & $-0.748$ & $-0.603$ & $-0.739$ & $-0.649$ \\
     & $\Theta$ & value & 0.578 & 0.548 & 0.582 & 0.545 \\
     &  & rank & 0.599 & 0.593 & 0.607 & 0.600 \\
     &  & concat & 0.586 & 0.554 & 0.584 & 0.553 \\
    \midrule
    RQ3 & G1 & value & 0.227 / 0.500 & 0.201 / 0.448 & 0.250 / 0.520 & 0.151 / 0.440 \\
     &  & rank & 0.621 / 0.623 & 0.575 / 0.573 & 0.612 / 0.605 & 0.569 / 0.569 \\
     &  & concat & 0.235 / 0.632 & 0.276 / 0.582 & 0.255 / 0.621 & 0.232 / 0.585 \\
     & G2 & value & 0.191 / 0.475 & 0.199 / 0.411 & 0.182 / 0.486 & 0.171 / 0.413 \\
     &  & rank & 0.586 / 0.581 & 0.580 / 0.587 & 0.584 / 0.570 & 0.572 / 0.574 \\
     &  & concat & 0.183 / 0.579 & 0.254 / 0.567 & 0.191 / 0.574 & 0.216 / 0.576 \\
     & G3 & value & $-0.630$ / 0.255 & $-0.709$ / 0.156 & $-0.655$ / 0.216 & $-0.804$ / 0.136 \\
     &  & rank & $-0.835$ / 0.156 & $-0.849$ / 0.134 & $-0.869$ / 0.113 & $-0.887$ / 0.109 \\
     &  & concat & $-0.624$ / 0.295 & $-0.681$ / 0.235 & $-0.621$ / 0.257 & $-0.769$ / 0.226 \\
     & $\Theta$ & value & 0.575 / 0.619 & 0.535 / 0.594 & 0.583 / 0.622 & 0.531 / 0.597 \\
     &  & rank & 0.599 / 0.630 & 0.593 / 0.626 & 0.607 / 0.628 & 0.600 / 0.624 \\
     &  & concat & 0.579 / 0.634 & 0.544 / 0.623 & 0.584 / 0.638 & 0.545 / 0.627 \\
    \bottomrule
  \end{tabular}
\end{table}

\FloatBarrier
\section{Generalization to Unseen Functions}
\label{app:lofo}
Our remedies address distribution shifts that keep the order of objective values (\cref{sec:discussion}); unseen functions are not such a shift.
We checked whether the remedies nevertheless improve the generalization of the regression of $J$ to unseen functions with leave-one-function-out (LOFO) cross-validation.
Each of the 24 folds trains on the training instances of 23 functions (2,300 instances) and predicts $J$ on the 10 untransformed test instances of the held-out function; data augmentation adds transformed copies of the 2,300 instances as in RQ3.
Conditions, learning models, seeds, and $\omega$ are the same as in RQ3.
We compute $R^2$ once per pair of model seed and $\omega$ over the pooled 240 predictions.
The LOFO $R^2$ was negative in all 36 conditions (\cref{tab:app-lofo}), with medians from $-0.94$ to $-0.16$; even the upper quartile over seeds was negative in every condition (at most $-0.10$).
The remedies changed the magnitude but not the sign, so none of them helped on unseen functions.

\begin{table}[!htbp]
  \centering
  \caption{LOFO $R^{2}$ is negative in every condition (median over 11 model seeds).}
  \label{tab:app-lofo}
  \small
  \begin{tabular}{@{}llrrrrrr@{}}
    \toprule
    & & \multicolumn{3}{c}{ELA} & \multicolumn{3}{c}{Deep-ELA} \\
    \cmidrule(lr){3-5}\cmidrule(lr){6-8}
    Extraction & Augmentation & G1 & G2 & G3 & G1 & G2 & G3 \\
    \midrule
    value  & no  & $-0.86$ & $-0.94$ & $-0.20$ & $-0.61$ & $-0.69$ & $-0.20$ \\
           & yes & $-0.81$ & $-0.67$ & $-0.34$ & $-0.51$ & $-0.58$ & $-0.38$ \\
    rank   & no  & $-0.70$ & $-0.65$ & $-0.32$ & $-0.69$ & $-0.65$ & $-0.38$ \\
           & yes & $-0.63$ & $-0.63$ & $-0.37$ & $-0.72$ & $-0.67$ & $-0.49$ \\
    concat & no  & $-0.89$ & $-0.90$ & $-0.20$ & $-0.57$ & $-0.63$ & $-0.16$ \\
           & yes & $-0.73$ & $-0.66$ & $-0.35$ & $-0.58$ & $-0.66$ & $-0.50$ \\
    \bottomrule
  \end{tabular}
\end{table}

\FloatBarrier
\section{Bootstrap Intervals}
\label{app:bootstrap}
The tests in the main paper cover only the variation across seeds of the learning model (\cref{subsec:exp-tests}).
This appendix gives percentile bootstrap intervals that also cover the seeds $\omega$ and the test instances.
Each of 10,000 bootstrap samples draws with replacement the 11 model seeds, the 11 seeds $\omega$, and, for each of the 24 functions, its 10 test instances.
For each sample, we recompute $R^{2}$ on the drawn instances and $\Theta$ as their mean.
As for the levels in the main paper, we then take the median over $\omega$ for each model seed and the median over the model seeds.
We take each difference within a sample.
For data augmentation, we draw the model seeds of the two training conditions independently, as in the Mann--Whitney $U$ test.
The interval runs from the 2.5th to the 97.5th percentile of the difference.

This procedure keeps the 24 functions fixed, as LIO evaluation does.
A second version also draws the 24 functions with replacement before it draws their instances.
It asks whether a difference would hold on other functions, which our experiments do not test (Appendix~\ref{app:lofo}).
\Cref{tab:app-bootstrap-ela,tab:app-bootstrap-deep-ela} list both intervals.

\paragraph{Functions fixed.}
With ELA, the interval excludes zero for the following differences.
\begin{itemize}
  \item RQ1: after the transformation, the baseline exceeded the predictor with all features by 0.219 ($[0.073, 0.343]$).
  \item RQ2: after the transformation, rank-based extraction exceeded the other two extractions for G1 and G2; before the transformation, it was below value-based extraction in every group.
  \item RQ3: data augmentation raised the $R^{2}$ of value-based extraction and concatenation in every group and $\Theta$ for every extraction; on the original test instances, it lowered the $R^{2}$ of value-based extraction and concatenation in every group.
\end{itemize}
The interval includes zero for the following differences.
\begin{itemize}
  \item RQ2: the advantage of rank-based extraction in the selection task (0.021, $[-0.005, 0.046]$) and the differences for G3 after the transformation.
  \item RQ3: the change of rank-based extraction for G1 and G2 with data augmentation.
  After data augmentation, the difference between concatenation and rank-based extraction for G1 (0.031, $[-0.014, 0.077]$) and G2, all differences for G3, and the differences in $\Theta$ except that between concatenation and value-based extraction.
\end{itemize}
With Deep-ELA, the intervals agree with the following exceptions.
The advantage of rank-based extraction in the selection task excludes zero (0.052, $[0.018, 0.084]$).
For G3 after the transformation, rank-based extraction is below the other two extractions.
After data augmentation, concatenation exceeds value-based extraction for G3, and rank-based extraction exceeds value-based extraction in $\Theta$.
The drawback of data augmentation on the original test instances includes zero for value-based extraction for G1 and G2.

\paragraph{Functions resampled.}
All intervals widen.
With ELA, the interval for RQ1 includes zero ($[-0.100, 0.621]$): the 24 functions differ too much to decide whether the baseline would exceed the predictor with all features on other functions.
The drop of the predictor with all features still excludes zero (0.619, $[0.352, 1.085]$), as do the advantage of rank-based extraction for G1 and G2 and the gain from data augmentation for value-based extraction and concatenation, but not this gain in $\Theta$ for rank-based extraction ($[{-}0.006, 0.088]$).
The main paper therefore states its direct evidence for the 24 BBOB functions under LIO evaluation (\cref{sec:discussion}).

\begin{table}[!htbp]
  \centering
  \caption{Bootstrap 95\% intervals of the differences with ELA. \emph{Estimate}: the difference of the levels in the main paper. \emph{Functions fixed}: resampling model seeds, $\omega$, and the test instances of each function. \emph{Functions resampled}: resampling the 24 functions as well. $\dagger$: the interval includes zero.}
  \label{tab:app-bootstrap-ela}
  \scriptsize
  \setlength{\tabcolsep}{4pt}
  \begin{tabular}{@{}llllrll@{}}
    \toprule
    & Task & Test instances & Difference & Estimate & Functions fixed & Functions resampled \\
    \midrule
    RQ1 & G1 & transformed & invariant $-$ all features & 0.219 & $[0.073, 0.343]$ & $[{-}0.100, 0.621]$$^{\dagger}$ \\
     & G1 & -- & original $-$ transformed, all features & 0.619 & $[0.486, 0.740]$ & $[0.352, 1.085]$ \\
     & G1 & -- & original $-$ transformed, invariant features & 0.000 & $[0.000, 0.000]$ & $[0.000, 0.000]$ \\
     & G1 & original & all $-$ invariant features & 0.400 & $[0.341, 0.469]$ & $[0.276, 0.617]$ \\
    \midrule
    RQ2 & G1 & transformed & rank $-$ value & 0.428 & $[0.287, 0.556]$ & $[0.144, 0.869]$ \\
     & G1 & transformed & rank $-$ concat & 0.407 & $[0.256, 0.559]$ & $[0.133, 0.848]$ \\
     & G1 & original & rank $-$ value & $-0.190$ & $[{-}0.242, {-}0.140]$ & $[{-}0.326, {-}0.102]$ \\
     & G2 & transformed & rank $-$ value & 0.411 & $[0.248, 0.537]$ & $[0.142, 0.704]$ \\
     & G2 & transformed & rank $-$ concat & 0.423 & $[0.260, 0.573]$ & $[0.135, 0.748]$ \\
     & G2 & original & rank $-$ value & $-0.148$ & $[{-}0.215, {-}0.093]$ & $[{-}0.283, {-}0.062]$ \\
     & G3 & transformed & rank $-$ value & $-0.151$ & $[{-}0.312, 0.005]$$^{\dagger}$ & $[{-}0.575, 0.295]$$^{\dagger}$ \\
     & G3 & transformed & rank $-$ concat & $-0.145$ & $[{-}0.320, 0.061]$$^{\dagger}$ & $[{-}0.558, 0.333]$$^{\dagger}$ \\
     & G3 & original & rank $-$ value & $-0.149$ & $[{-}0.205, {-}0.108]$ & $[{-}0.319, {-}0.056]$ \\
     & G1 & -- & original $-$ transformed, rank & 0.000 & $[0.000, 0.000]$ & $[0.000, 0.000]$ \\
     & G1 & -- & original $-$ transformed, concat & 0.595 & $[0.451, 0.743]$ & $[0.340, 1.059]$ \\
     & G2 & -- & original $-$ transformed, value & 0.574 & $[0.418, 0.712]$ & $[0.340, 0.895]$ \\
     & G2 & -- & original $-$ transformed, rank & 0.015 & $[{-}0.022, 0.063]$$^{\dagger}$ & $[{-}0.042, 0.088]$$^{\dagger}$ \\
     & G2 & -- & original $-$ transformed, concat & 0.575 & $[0.427, 0.736]$ & $[0.331, 0.935]$ \\
     & $\Theta$ & transformed & rank $-$ value & 0.021 & $[{-}0.005, 0.046]$$^{\dagger}$ & $[{-}0.044, 0.086]$$^{\dagger}$ \\
     & $\Theta$ & transformed & rank $-$ concat & 0.020 & $[{-}0.008, 0.042]$$^{\dagger}$ & $[{-}0.047, 0.082]$$^{\dagger}$ \\
    \midrule
    RQ3 & G1 & transformed & augmented $-$ plain, value & 0.337 & $[0.271, 0.412]$ & $[0.154, 0.648]$ \\
     & G1 & transformed & augmented $-$ plain, rank & $-0.006$ & $[{-}0.044, 0.037]$$^{\dagger}$ & $[{-}0.056, 0.053]$$^{\dagger}$ \\
     & G1 & transformed & augmented $-$ plain, concat & 0.425 & $[0.343, 0.534]$ & $[0.240, 0.759]$ \\
     & G2 & transformed & augmented $-$ plain, value & 0.273 & $[0.191, 0.320]$ & $[0.090, 0.453]$ \\
     & G2 & transformed & augmented $-$ plain, rank & 0.003 & $[{-}0.026, 0.072]$$^{\dagger}$ & $[{-}0.045, 0.099]$$^{\dagger}$ \\
     & G2 & transformed & augmented $-$ plain, concat & 0.333 & $[0.251, 0.408]$ & $[0.166, 0.544]$ \\
     & G3 & transformed & augmented $-$ plain, value & 0.970 & $[0.831, 1.090]$ & $[0.368, 2.269]$ \\
     & G3 & transformed & augmented $-$ plain, rank & 1.170 & $[0.995, 1.292]$ & $[0.438, 2.582]$ \\
     & G3 & transformed & augmented $-$ plain, concat & 0.965 & $[0.849, 1.111]$ & $[0.391, 2.310]$ \\
     & G1 & original & augmented $-$ plain, value & $-0.106$ & $[{-}0.146, {-}0.059]$ & $[{-}0.193, {-}0.033]$ \\
     & G1 & original & augmented $-$ plain, rank & $-0.006$ & $[{-}0.044, 0.037]$$^{\dagger}$ & $[{-}0.056, 0.053]$$^{\dagger}$ \\
     & G1 & original & augmented $-$ plain, concat & $-0.086$ & $[{-}0.130, {-}0.042]$ & $[{-}0.176, {-}0.021]$ \\
     & G2 & original & augmented $-$ plain, value & $-0.086$ & $[{-}0.138, {-}0.044]$ & $[{-}0.192, {-}0.015]$ \\
     & G2 & original & augmented $-$ plain, rank & 0.011 & $[{-}0.029, 0.071]$$^{\dagger}$ & $[{-}0.048, 0.099]$$^{\dagger}$ \\
     & G2 & original & augmented $-$ plain, concat & $-0.056$ & $[{-}0.105, {-}0.012]$ & $[{-}0.144, 0.007]$$^{\dagger}$ \\
     & G3 & original & augmented $-$ plain, value & $-0.074$ & $[{-}0.120, {-}0.037]$ & $[{-}0.162, {-}0.014]$ \\
     & G3 & original & augmented $-$ plain, rank & $-0.307$ & $[{-}0.354, {-}0.259]$ & $[{-}0.557, {-}0.085]$ \\
     & G3 & original & augmented $-$ plain, concat & $-0.064$ & $[{-}0.108, {-}0.030]$ & $[{-}0.139, {-}0.008]$ \\
     & $\Theta$ & transformed & augmented $-$ plain, value & 0.058 & $[0.046, 0.068]$ & $[0.020, 0.101]$ \\
     & $\Theta$ & transformed & augmented $-$ plain, rank & 0.027 & $[0.016, 0.045]$ & $[{-}0.006, 0.088]$$^{\dagger}$ \\
     & $\Theta$ & transformed & augmented $-$ plain, concat & 0.054 & $[0.045, 0.070]$ & $[0.019, 0.104]$ \\
     & G1 & transformed & concat $-$ rank, augmented & 0.031 & $[{-}0.014, 0.077]$$^{\dagger}$ & $[{-}0.043, 0.109]$$^{\dagger}$ \\
     & G1 & transformed & concat $-$ value, augmented & 0.104 & $[0.057, 0.148]$ & $[0.033, 0.196]$ \\
     & G1 & transformed & rank $-$ value, augmented & 0.073 & $[0.014, 0.131]$ & $[{-}0.019, 0.182]$$^{\dagger}$ \\
     & G2 & transformed & concat $-$ rank, augmented & $-0.004$ & $[{-}0.061, 0.031]$$^{\dagger}$ & $[{-}0.099, 0.063]$$^{\dagger}$ \\
     & G2 & transformed & concat $-$ value, augmented & 0.087 & $[0.042, 0.135]$ & $[0.025, 0.175]$ \\
     & G2 & transformed & rank $-$ value, augmented & 0.091 & $[0.050, 0.152]$ & $[0.008, 0.217]$ \\
     & G3 & transformed & concat $-$ rank, augmented & 0.018 & $[{-}0.066, 0.133]$$^{\dagger}$ & $[{-}0.118, 0.232]$$^{\dagger}$ \\
     & G3 & transformed & concat $-$ value, augmented & $-0.024$ & $[{-}0.062, 0.071]$$^{\dagger}$ & $[{-}0.089, 0.098]$$^{\dagger}$ \\
     & G3 & transformed & rank $-$ value, augmented & $-0.042$ & $[{-}0.131, 0.072]$$^{\dagger}$ & $[{-}0.229, 0.134]$$^{\dagger}$ \\
     & $\Theta$ & transformed & concat $-$ rank, augmented & 0.006 & $[{-}0.005, 0.020]$$^{\dagger}$ & $[{-}0.024, 0.042]$$^{\dagger}$ \\
     & $\Theta$ & transformed & concat $-$ value, augmented & 0.010 & $[0.004, 0.022]$ & $[{-}0.001, 0.030]$$^{\dagger}$ \\
     & $\Theta$ & transformed & rank $-$ value, augmented & 0.004 & $[{-}0.008, 0.019]$$^{\dagger}$ & $[{-}0.034, 0.043]$$^{\dagger}$ \\
    \bottomrule
  \end{tabular}
\end{table}

\begin{table}[!htbp]
  \centering
  \caption{Bootstrap 95\% intervals of the differences with Deep-ELA. \emph{Estimate}: the difference of the levels in the main paper. \emph{Functions fixed}: resampling model seeds, $\omega$, and the test instances of each function. \emph{Functions resampled}: resampling the 24 functions as well. $\dagger$: the interval includes zero.}
  \label{tab:app-bootstrap-deep-ela}
  \scriptsize
  \setlength{\tabcolsep}{4pt}
  \begin{tabular}{@{}llllrll@{}}
    \toprule
    & Task & Test instances & Difference & Estimate & Functions fixed & Functions resampled \\
    \midrule
    RQ2 & G1 & transformed & rank $-$ value & 0.484 & $[0.335, 0.620]$ & $[0.199, 0.912]$ \\
     & G1 & transformed & rank $-$ concat & 0.480 & $[0.317, 0.606]$ & $[0.191, 0.885]$ \\
     & G1 & original & rank $-$ value & $-0.230$ & $[{-}0.302, {-}0.165]$ & $[{-}0.411, {-}0.122]$ \\
     & G2 & transformed & rank $-$ value & 0.686 & $[0.526, 0.856]$ & $[0.341, 1.170]$ \\
     & G2 & transformed & rank $-$ concat & 0.593 & $[0.433, 0.758]$ & $[0.227, 1.070]$ \\
     & G2 & original & rank $-$ value & $-0.210$ & $[{-}0.284, {-}0.147]$ & $[{-}0.381, {-}0.094]$ \\
     & G3 & transformed & rank $-$ value & $-0.204$ & $[{-}0.450, {-}0.013]$ & $[{-}0.957, 0.371]$$^{\dagger}$ \\
     & G3 & transformed & rank $-$ concat & $-0.256$ & $[{-}0.467, {-}0.083]$ & $[{-}0.905, 0.195]$$^{\dagger}$ \\
     & G3 & original & rank $-$ value & $-0.205$ & $[{-}0.249, {-}0.147]$ & $[{-}0.366, {-}0.095]$ \\
     & G1 & -- & original $-$ transformed, rank & 0.000 & $[0.000, 0.000]$ & $[0.000, 0.000]$ \\
     & G1 & -- & original $-$ transformed, concat & 0.698 & $[0.554, 0.826]$ & $[0.445, 1.125]$ \\
     & G2 & -- & original $-$ transformed, value & 0.910 & $[0.761, 1.081]$ & $[0.646, 1.377]$ \\
     & G2 & -- & original $-$ transformed, rank & 0.014 & $[{-}0.025, 0.056]$$^{\dagger}$ & $[{-}0.044, 0.079]$$^{\dagger}$ \\
     & G2 & -- & original $-$ transformed, concat & 0.821 & $[0.668, 0.986]$ & $[0.538, 1.287]$ \\
     & $\Theta$ & transformed & rank $-$ value & 0.052 & $[0.018, 0.084]$ & $[{-}0.034, 0.128]$$^{\dagger}$ \\
     & $\Theta$ & transformed & rank $-$ concat & 0.046 & $[0.016, 0.080]$ & $[{-}0.029, 0.118]$$^{\dagger}$ \\
    \midrule
    RQ3 & G1 & transformed & augmented $-$ plain, value & 0.350 & $[0.282, 0.422]$ & $[0.165, 0.646]$ \\
     & G1 & transformed & augmented $-$ plain, rank & $-0.003$ & $[{-}0.038, 0.033]$$^{\dagger}$ & $[{-}0.054, 0.048]$$^{\dagger}$ \\
     & G1 & transformed & augmented $-$ plain, concat & 0.426 & $[0.367, 0.506]$ & $[0.259, 0.743]$ \\
     & G2 & transformed & augmented $-$ plain, value & 0.385 & $[0.317, 0.462]$ & $[0.218, 0.641]$ \\
     & G2 & transformed & augmented $-$ plain, rank & 0.010 & $[{-}0.025, 0.053]$$^{\dagger}$ & $[{-}0.045, 0.078]$$^{\dagger}$ \\
     & G2 & transformed & augmented $-$ plain, concat & 0.428 & $[0.351, 0.508]$ & $[0.221, 0.735]$ \\
     & G3 & transformed & augmented $-$ plain, value & 0.923 & $[0.810, 1.061]$ & $[0.482, 1.952]$ \\
     & G3 & transformed & augmented $-$ plain, rank & 1.115 & $[1.003, 1.287]$ & $[0.475, 2.459]$ \\
     & G3 & transformed & augmented $-$ plain, concat & 0.940 & $[0.823, 1.073]$ & $[0.466, 2.026]$ \\
     & G1 & original & augmented $-$ plain, value & $-0.028$ & $[{-}0.065, 0.008]$$^{\dagger}$ & $[{-}0.083, 0.021]$$^{\dagger}$ \\
     & G1 & original & augmented $-$ plain, rank & $-0.003$ & $[{-}0.038, 0.033]$$^{\dagger}$ & $[{-}0.054, 0.048]$$^{\dagger}$ \\
     & G1 & original & augmented $-$ plain, concat & $-0.041$ & $[{-}0.094, {-}0.006]$ & $[{-}0.123, 0.007]$$^{\dagger}$ \\
     & G2 & original & augmented $-$ plain, value & $-0.032$ & $[{-}0.079, 0.005]$$^{\dagger}$ & $[{-}0.110, 0.019]$$^{\dagger}$ \\
     & G2 & original & augmented $-$ plain, rank & 0.008 & $[{-}0.028, 0.055]$$^{\dagger}$ & $[{-}0.048, 0.078]$$^{\dagger}$ \\
     & G2 & original & augmented $-$ plain, concat & $-0.061$ & $[{-}0.109, {-}0.015]$ & $[{-}0.147, 0.003]$$^{\dagger}$ \\
     & G3 & original & augmented $-$ plain, value & $-0.099$ & $[{-}0.144, {-}0.065]$ & $[{-}0.215, {-}0.023]$ \\
     & G3 & original & augmented $-$ plain, rank & $-0.292$ & $[{-}0.352, {-}0.252]$ & $[{-}0.560, {-}0.062]$ \\
     & G3 & original & augmented $-$ plain, concat & $-0.116$ & $[{-}0.157, {-}0.073]$ & $[{-}0.234, {-}0.035]$ \\
     & $\Theta$ & transformed & augmented $-$ plain, value & 0.063 & $[0.048, 0.077]$ & $[0.022, 0.108]$ \\
     & $\Theta$ & transformed & augmented $-$ plain, rank & 0.030 & $[0.011, 0.042]$ & $[{-}0.013, 0.098]$$^{\dagger}$ \\
     & $\Theta$ & transformed & augmented $-$ plain, concat & 0.084 & $[0.070, 0.101]$ & $[0.043, 0.132]$ \\
     & G1 & transformed & concat $-$ rank, augmented & 0.002 & $[{-}0.039, 0.056]$$^{\dagger}$ & $[{-}0.066, 0.093]$$^{\dagger}$ \\
     & G1 & transformed & concat $-$ value, augmented & 0.172 & $[0.130, 0.230]$ & $[0.099, 0.297]$ \\
     & G1 & transformed & rank $-$ value, augmented & 0.171 & $[0.106, 0.239]$ & $[0.052, 0.319]$ \\
     & G2 & transformed & concat $-$ rank, augmented & 0.006 & $[{-}0.045, 0.055]$$^{\dagger}$ & $[{-}0.080, 0.101]$$^{\dagger}$ \\
     & G2 & transformed & concat $-$ value, augmented & 0.164 & $[0.111, 0.229]$ & $[0.083, 0.281]$ \\
     & G2 & transformed & rank $-$ value, augmented & 0.157 & $[0.086, 0.246]$ & $[0.029, 0.314]$ \\
     & G3 & transformed & concat $-$ rank, augmented & 0.074 & $[{-}0.025, 0.154]$$^{\dagger}$ & $[{-}0.172, 0.306]$$^{\dagger}$ \\
     & G3 & transformed & concat $-$ value, augmented & 0.074 & $[0.017, 0.120]$ & $[{-}0.020, 0.166]$$^{\dagger}$ \\
     & G3 & transformed & rank $-$ value, augmented & 0.000 & $[{-}0.089, 0.098]$$^{\dagger}$ & $[{-}0.258, 0.261]$$^{\dagger}$ \\
     & $\Theta$ & transformed & concat $-$ rank, augmented & 0.004 & $[{-}0.008, 0.021]$$^{\dagger}$ & $[{-}0.026, 0.041]$$^{\dagger}$ \\
     & $\Theta$ & transformed & concat $-$ value, augmented & 0.039 & $[0.029, 0.051]$ & $[0.020, 0.063]$ \\
     & $\Theta$ & transformed & rank $-$ value, augmented & 0.036 & $[0.016, 0.051]$ & $[{-}0.012, 0.079]$$^{\dagger}$ \\
    \bottomrule
  \end{tabular}
\end{table}

\FloatBarrier
\section{Additional Controls}
\label{app:controls}
This appendix reports controls for statements in the main paper.
They add test sets and one training condition; all other settings are the same as in the main paper.
We reran all algorithms on every new instance.

\paragraph{The rescaling alone.}
The transformation $g_{\psi,p}$ rescales the objective values to $[0,1]$ and then applies $\psi$ (\cref{eq:warp}).
To separate the two steps, we created a transformed test set with $\psi$ set to the identity, so that only a positive affine transformation remains (\emph{Identity} in \cref{tab:app-blocks}).
The features and $J$ of G1 did not change, so its $R^{2}$ did not change.
For G2, $J$ changed by 0.09 (0.25 with $\psi\sim\Dspline$), and the $R^{2}$ of value-based extraction fell only from 0.77 to 0.76.
For G3, $J$ changed by 0.65 (0.80 with $\psi\sim\Dspline$), and the $R^{2}$ of value-based extraction fell from 0.78 to $-0.72$.
Dual Annealing accepts solutions based on differences in objective value, so the rescaling changes its search, whereas min-max normalization and ranks remove the scale from the features.
On this test set, the input of every feature extraction is the same as on the original test instances, so no model can follow the change in $J$.
With data augmentation, the $R^{2}$ for G3 therefore remained negative ($-0.43$ for value-based extraction, $-0.01$ for rank-based extraction, and $-0.51$ for concatenation), whereas it was 0.18--0.34 with $\psi\sim\Dspline$.
The gain from data augmentation for G3 in \cref{sec:rq3} thus depends on $\psi$ accompanying the rescaling in both the training and the test instances: the change in the input caused by $\psi$ indicates that $J$ has changed.
The order of the feature extractions also differs: with the rescaling alone, rank-based extraction was the highest after data augmentation.
For rank-based extraction, an instance and its transformed copy have the same input but, for G3, a different $J$, so a model trained with data augmentation cannot fit both; its $R^{2}$ on the original test instances fell from 0.63 to 0.33 (\cref{tab:app-results}).

\paragraph{Further draws of $\psi$.}
The main paper uses one draw of $\psi_{p}\sim\Dspline$ for each test instance.
We repeated the \testname{} with four more draws (\emph{Draws 1--4} in \cref{tab:app-blocks}).
The conclusions did not change.
For G1, the $R^{2}$ of the predictor with all features was between $-0.04$ and 0.13, below the 0.23 in the main paper and below the 0.45 of the baseline.
The $R^{2}$ of rank-based extraction for G1 and G2 was at least 0.61, and data augmentation raised the $R^{2}$ of value-based extraction and concatenation in every group and every draw.

\paragraph{Training-set size.}
Data augmentation doubles the training set.
To separate this effect from that of the transformation, we trained models on the 2,400 training instances and 2,400 new untransformed instances (100 more instances of each function), with the same number of parameter updates as with data augmentation (\emph{plain-4800} in \cref{tab:app-doubled}).
The SBS is the same in all training conditions.
On the transformed test instances, plain-4800 stayed near the level without data augmentation for value-based extraction and concatenation; with ELA, for example, the $R^{2}$ of value-based extraction for G1 was 0.25 without data augmentation, 0.23 with plain-4800, and 0.58 with data augmentation.
This holds for each of the four distributions (\cref{tab:app-families}).
The larger training set therefore does not explain the gain from data augmentation.
On the original test instances, plain-4800 gave the highest $R^{2}$ and $\Theta$ in every cell.
With the same number of additional algorithm runs, new instances thus raise prediction performance on the original instances, and transformed copies raise it on the transformed instances.
For rank-based extraction for G1 and G2, plain-4800 was also the highest on the transformed instances, because the transformed copies add little to rank-based features.
With ELA, its $R^{2}$ there (0.74 for G1 and 0.71 for G2) was also above that of every feature extraction with data augmentation (at most 0.69 and 0.62); we did not test differences across both feature extractions and training conditions.
In the selection task, in contrast, data augmentation gave the highest $\Theta$ on the transformed instances (0.65, versus at most 0.63 with plain-4800).
All differences between data augmentation and plain-4800 were significant (Mann--Whitney $U$ test with Holm correction, $p \le 0.001$).

\paragraph{Results of RQ3 on each distribution.}
The main paper reports the mean over the four distributions that data augmentation does not use.
Data augmentation raised the $R^{2}$ of value-based extraction and concatenation for every group, and $\Theta$ for every feature extraction, on each of the four distributions, with ELA (\cref{tab:app-families}) and with Deep-ELA.

\paragraph{Drawback of data augmentation in the selection task.}
On the original test instances, data augmentation lowered the $\Theta$ of value-based extraction from 0.704 to 0.681 and that of concatenation from 0.704 to 0.683 with ELA (both significant), and raised that of rank-based extraction from 0.646 to 0.650 (\cref{tab:app-doubled}).
With Deep-ELA, it lowered the $\Theta$ of concatenation from 0.708 to 0.695 (significant); the other two changed by at most 0.004 (not significant).

\paragraph{Change in the prediction for G1.}
For G1, the transformation does not change $J$, so a change in the prediction measures the reliance on information that the algorithms ignore.
We computed the mean of $\lvert \hat{J}_{\tau}(\tilde{p},a) - \hat{J}_{\tau}(p,a) \rvert$ over the test instances, on the four distributions that data augmentation does not use.
With ELA, data augmentation reduced it from 0.189 to 0.086 for value-based extraction and from 0.186 to 0.053 for concatenation; with Deep-ELA, from 0.227 to 0.146 and from 0.205 to 0.083 (all significant).
It is zero for rank-based extraction.

\paragraph{Variance of $J$ and prediction error.}
A lower $R^{2}$ can also come from a smaller variance of $J$.
The variance of $J$ over the test instances was 0.154 for G1 and 0.067 for G2 before and after the transformation by $\Dspline$, and fell from 0.119 to 0.097 for G3.
Without data augmentation, the mean squared error of value-based extraction with ELA rose from 0.023 to 0.118 for G1, from 0.015 to 0.054 for G2, and from 0.026 to 0.174 for G3.
The drops in $R^{2}$ therefore come from larger errors.

\paragraph{Identity of the runs for G1.}
In the regression task, every run of the two G1 algorithms gave the same result on an instance and on its transformed copy: none of the 148,800 runs on the training instances and none of the 14,880 runs on each of the ten transformed test sets differed in the best objective value, the number of evaluations, or the number of restarts.
With the algorithm settings of the selection task (\cref{tab:app-role-selected}), up to 2.5\% of the runs of Adaptive Nelder--Mead and 0.5\% of those of BIPOP CMA-ES on a transformed test set differed in the best objective value or the number of restarts, but never in the number of evaluations.
We attribute these differences to floating-point rounding, because a transformation can map two close objective values to the same number.

\begin{table}[!htbp]
  \centering
  \caption{Results on further transformed test sets. \emph{Identity}: $\psi$ is the identity, so only the rescaling in \cref{eq:warp} remains. \emph{Main}: the draw of $\psi\sim\Dspline$ in the main paper. \emph{Draws 1--4}: four more draws. $\Delta_{J}$ is the change in $J$ (\cref{eq:variation}); $R^{2}$ and $\Theta$ are medians over 11 model seeds, without (plain) and with (aug.) data augmentation.}
  \label{tab:app-blocks}
  \footnotesize
  \setlength{\tabcolsep}{5pt}
  \begin{tabular}{@{}lllrrrrrrr@{}}
    \toprule
    & & & Original & Identity & Main & Draw 1 & Draw 2 & Draw 3 & Draw 4 \\
    \midrule
    $\Delta_{J}$ & G1 & & -- & 0.000 & 0.000 & 0.000 & 0.000 & 0.000 & 0.000 \\
     & G2 & & -- & 0.085 & 0.249 & 0.218 & 0.224 & 0.221 & 0.216 \\
     & G3 & & -- & 0.654 & 0.802 & 0.784 & 0.769 & 0.772 & 0.785 \\
    $J$ of BOBYQA & G3 & & 0.513 & 0.719 & 0.687 & 0.681 & 0.679 & 0.683 & 0.677 \\
    \midrule
    $R^{2}$ & G1 & value, plain & 0.852 & 0.852 & 0.233 & $-0.038$ & 0.129 & 0.055 & 0.041 \\
     &  & value, aug. & 0.746 & 0.746 & 0.610 & 0.545 & 0.598 & 0.594 & 0.547 \\
     &  & rank, plain & 0.661 & 0.661 & 0.661 & 0.661 & 0.661 & 0.661 & 0.661 \\
     &  & rank, aug. & 0.655 & 0.655 & 0.655 & 0.655 & 0.655 & 0.655 & 0.655 \\
     &  & concat, plain & 0.849 & 0.849 & 0.255 & $-0.049$ & 0.156 & 0.045 & 0.062 \\
     &  & concat, aug. & 0.764 & 0.764 & 0.691 & 0.660 & 0.687 & 0.698 & 0.663 \\
    & & invariant, plain & 0.452 & 0.452 & 0.452 & 0.452 & 0.452 & 0.452 & 0.452 \\
     & G2 & value, plain & 0.770 & 0.755 & 0.196 & 0.089 & 0.068 & 0.111 & 0.105 \\
     &  & value, aug. & 0.684 & 0.673 & 0.545 & 0.485 & 0.527 & 0.538 & 0.488 \\
     &  & rank, plain & 0.621 & 0.603 & 0.607 & 0.629 & 0.622 & 0.619 & 0.623 \\
     &  & rank, aug. & 0.632 & 0.620 & 0.599 & 0.626 & 0.633 & 0.623 & 0.629 \\
     &  & concat, plain & 0.759 & 0.758 & 0.184 & 0.093 & 0.056 & 0.109 & 0.090 \\
     &  & concat, aug. & 0.703 & 0.693 & 0.632 & 0.602 & 0.627 & 0.627 & 0.599 \\
     & G3 & value, plain & 0.784 & $-0.723$ & $-0.798$ & $-0.769$ & $-0.874$ & $-0.735$ & $-0.736$ \\
     &  & value, aug. & 0.710 & $-0.432$ & 0.273 & 0.218 & 0.374 & 0.336 & 0.261 \\
     &  & rank, plain & 0.635 & $-0.868$ & $-0.949$ & $-0.943$ & $-0.924$ & $-0.954$ & $-0.902$ \\
     &  & rank, aug. & 0.328 & $-0.014$ & 0.182 & 0.191 & 0.189 & 0.167 & 0.184 \\
     &  & concat, plain & 0.782 & $-0.737$ & $-0.804$ & $-0.759$ & $-0.878$ & $-0.699$ & $-0.766$ \\
     &  & concat, aug. & 0.718 & $-0.512$ & 0.335 & 0.280 & 0.393 & 0.337 & 0.317 \\
    \midrule
    $\Theta$ & & value, plain & 0.704 & 0.704 & 0.598 & 0.574 & 0.575 & 0.589 & 0.564 \\
     & & value, aug. & 0.681 & 0.681 & 0.650 & 0.640 & 0.647 & 0.652 & 0.640 \\
     & & rank, plain & 0.646 & 0.645 & 0.620 & 0.618 & 0.621 & 0.621 & 0.620 \\
     & & rank, aug. & 0.650 & 0.650 & 0.647 & 0.647 & 0.648 & 0.649 & 0.648 \\
     & & concat, plain & 0.704 & 0.704 & 0.600 & 0.577 & 0.586 & 0.600 & 0.570 \\
     & & concat, aug. & 0.683 & 0.683 & 0.654 & 0.652 & 0.656 & 0.662 & 0.656 \\
    \bottomrule
  \end{tabular}
\end{table}

\begin{table}[!htbp]
  \centering
  \caption{Data augmentation against a training set of the same size without transformations (medians over 11 model seeds). \emph{Plain}: 2,400 training instances. \emph{Plain-4800}: these and 2,400 new untransformed instances. \emph{Aug.}: these and their 2,400 transformed copies. \emph{Transformed}: mean over the four distributions that data augmentation does not use. Bold marks the best training condition in each cell.}
  \label{tab:app-doubled}
  \footnotesize
  \begin{tabular}{@{}lllrrrrrr@{}}
    \toprule
    & & & \multicolumn{3}{c}{Original test instances} & \multicolumn{3}{c}{Transformed test instances} \\
    \cmidrule(lr){4-6}\cmidrule(lr){7-9}
    & Task & Extraction & Plain & Plain-4800 & Aug. & Plain & Plain-4800 & Aug. \\
    \midrule
    ELA & G1 ($R^{2}$) & value & 0.852 & \textbf{0.895} & 0.746 & 0.245 & 0.229 & \textbf{0.582} \\
     &  & rank & 0.661 & \textbf{0.738} & 0.655 & 0.661 & \textbf{0.738} & 0.655 \\
     &  & concat & 0.849 & \textbf{0.885} & 0.764 & 0.262 & 0.245 & \textbf{0.686} \\
     & G2 ($R^{2}$) & value & 0.770 & \textbf{0.829} & 0.684 & 0.252 & 0.266 & \textbf{0.525} \\
     &  & rank & 0.621 & \textbf{0.735} & 0.632 & 0.613 & \textbf{0.713} & 0.616 \\
     &  & concat & 0.759 & \textbf{0.830} & 0.703 & 0.280 & 0.269 & \textbf{0.613} \\
     & G3 ($R^{2}$) & value & 0.784 & \textbf{0.839} & 0.710 & $-0.814$ & $-0.775$ & \textbf{0.157} \\
     &  & rank & 0.635 & \textbf{0.729} & 0.328 & $-1.055$ & $-0.987$ & \textbf{0.115} \\
     &  & concat & 0.782 & \textbf{0.841} & 0.718 & $-0.832$ & $-0.799$ & \textbf{0.133} \\
    & Mixed ($\Theta$) & value & 0.704 & \textbf{0.720} & 0.681 & 0.583 & 0.589 & \textbf{0.642} \\
    &  & rank & 0.646 & \textbf{0.666} & 0.650 & 0.619 & 0.633 & \textbf{0.646} \\
    &  & concat & 0.704 & \textbf{0.719} & 0.683 & 0.598 & 0.594 & \textbf{0.652} \\
    \midrule
    Deep-ELA & G1 ($R^{2}$) & value & 0.845 & \textbf{0.875} & 0.817 & 0.091 & 0.062 & \textbf{0.441} \\
     &  & rank & 0.615 & \textbf{0.667} & 0.612 & 0.615 & \textbf{0.667} & 0.612 \\
     &  & concat & 0.834 & \textbf{0.879} & 0.792 & 0.187 & 0.178 & \textbf{0.613} \\
     & G2 ($R^{2}$) & value & 0.794 & \textbf{0.825} & 0.762 & 0.036 & 0.022 & \textbf{0.421} \\
     &  & rank & 0.584 & \textbf{0.664} & 0.592 & 0.568 & \textbf{0.642} & 0.578 \\
     &  & concat & 0.798 & \textbf{0.828} & 0.737 & 0.157 & 0.184 & \textbf{0.585} \\
     & G3 ($R^{2}$) & value & 0.825 & \textbf{0.843} & 0.727 & $-0.800$ & $-0.923$ & \textbf{0.123} \\
     &  & rank & 0.620 & \textbf{0.686} & 0.328 & $-0.992$ & $-0.966$ & \textbf{0.123} \\
     &  & concat & 0.820 & \textbf{0.846} & 0.703 & $-0.743$ & $-0.811$ & \textbf{0.197} \\
    & Mixed ($\Theta$) & value & 0.707 & \textbf{0.715} & 0.703 & 0.543 & 0.544 & \textbf{0.606} \\
    &  & rank & 0.643 & \textbf{0.657} & 0.645 & 0.611 & 0.620 & \textbf{0.642} \\
    &  & concat & 0.708 & \textbf{0.715} & 0.695 & 0.561 & 0.564 & \textbf{0.645} \\
    \bottomrule
  \end{tabular}
\end{table}

\begin{table}[!htbp]
  \centering
  \caption{Results of RQ3 with ELA on each of the four distributions that data augmentation does not use (medians over 11 model seeds). Training conditions as in \cref{tab:app-doubled}. The main paper reports the mean over the four distributions; for G3, some values stay negative after data augmentation.}
  \label{tab:app-families}
  \footnotesize
  \begin{tabular}{@{}lllrrrr@{}}
    \toprule
    Task & Extraction & Training & $\Dspline$, $K\in\{2,3\}$ & $\Dspline$, $K\in\{14,\dots,19\}$ & Power & Piecewise linear \\
    \midrule
    G1 ($R^{2}$) & value & plain & 0.311 & 0.150 & 0.297 & 0.191 \\
     &  & plain-4800 & 0.307 & 0.074 & 0.302 & 0.159 \\
     &  & aug. & 0.609 & 0.594 & 0.547 & 0.590 \\
     & rank & plain & 0.661 & 0.661 & 0.661 & 0.661 \\
     &  & plain-4800 & 0.738 & 0.738 & 0.738 & 0.738 \\
     &  & aug. & 0.655 & 0.655 & 0.655 & 0.655 \\
     & concat & plain & 0.341 & 0.154 & 0.311 & 0.213 \\
     &  & plain-4800 & 0.368 & 0.117 & 0.325 & 0.190 \\
     &  & aug. & 0.690 & 0.702 & 0.647 & 0.696 \\
    \midrule
    G2 ($R^{2}$) & value & plain & 0.369 & 0.266 & 0.179 & 0.258 \\
     &  & plain-4800 & 0.413 & 0.236 & 0.190 & 0.264 \\
     &  & aug. & 0.541 & 0.561 & 0.429 & 0.554 \\
     & rank & plain & 0.609 & 0.605 & 0.606 & 0.624 \\
     &  & plain-4800 & 0.719 & 0.708 & 0.706 & 0.720 \\
     &  & aug. & 0.623 & 0.609 & 0.612 & 0.625 \\
     & concat & plain & 0.388 & 0.300 & 0.230 & 0.240 \\
     &  & plain-4800 & 0.383 & 0.205 & 0.216 & 0.246 \\
     &  & aug. & 0.628 & 0.641 & 0.526 & 0.643 \\
    \midrule
    G3 ($R^{2}$) & value & plain & $-0.841$ & $-1.004$ & $-0.738$ & $-0.723$ \\
     &  & plain-4800 & $-0.821$ & $-0.986$ & $-0.660$ & $-0.645$ \\
     &  & aug. & 0.035 & 0.353 & $-0.063$ & 0.239 \\
     & rank & plain & $-0.945$ & $-1.037$ & $-1.313$ & $-0.923$ \\
     &  & plain-4800 & $-0.893$ & $-0.992$ & $-1.230$ & $-0.845$ \\
     &  & aug. & 0.183 & 0.164 & $-0.031$ & 0.143 \\
     & concat & plain & $-0.857$ & $-0.995$ & $-0.772$ & $-0.705$ \\
     &  & plain-4800 & $-0.865$ & $-1.010$ & $-0.703$ & $-0.647$ \\
     &  & aug. & $-0.011$ & 0.366 & $-0.070$ & 0.224 \\
    \midrule
    Mixed ($\Theta$) & value & plain & 0.589 & 0.593 & 0.554 & 0.602 \\
     &  & plain-4800 & 0.592 & 0.592 & 0.560 & 0.610 \\
     &  & aug. & 0.647 & 0.647 & 0.625 & 0.647 \\
     & rank & plain & 0.621 & 0.616 & 0.613 & 0.624 \\
     &  & plain-4800 & 0.635 & 0.629 & 0.630 & 0.639 \\
     &  & aug. & 0.649 & 0.644 & 0.644 & 0.648 \\
     & concat & plain & 0.604 & 0.598 & 0.576 & 0.615 \\
     &  & plain-4800 & 0.599 & 0.594 & 0.564 & 0.613 \\
     &  & aug. & 0.657 & 0.658 & 0.641 & 0.662 \\
    \bottomrule
  \end{tabular}
\end{table}

\FloatBarrier

\end{document}